\pdfoutput=1

\documentclass[11pt]{article}

\usepackage[]{acl}
\usepackage{times}
\usepackage{latexsym}

\usepackage[T1]{fontenc}

\usepackage[utf8]{inputenc}

\usepackage{microtype}
\usepackage{inconsolata}
\usepackage{algorithm}
\usepackage{algorithmic}
\usepackage{multirow}
\usepackage{multicol}
\usepackage{booktabs}
\usepackage{marvosym}
\usepackage{colortbl}
\usepackage{arydshln}
\usepackage{pgfplots}
\usepackage{subfigure}
\usepackage{amsmath}
\usepackage{amsfonts}
\usepackage{pgf-pie} 
\usepackage{amssymb}
\title{Mars: Semantic-aware Contrastive Learning for \\ End-to-End Task-Oriented Dialog}
\title{Contrast Context & State Representations \\for End-to-End Task-Oriented Dialog}
\title{Improve End-to-End Task-Oriented Dialog via Contrasting Context & State Representations}
\title{Probing Context-and-State Representations with Contrastive Learning for End-to-End Task-Oriented Dialog}
\title{Pars: Probing Context \& State Representations with Contrastive Learning \\for End-to-End Task-Oriented Dialog}
\title{Mars: \underline{M}odeling Context \& St\underline{a}te \underline{R}epresentations with Contra\underline{s}tive Learning for End-to-End Task-Oriented Dialog}

\author{ Haipeng Sun, Junwei Bao\thanks{\;\;Corresponding author: baojunwei001@gmail.com}, Youzheng Wu, Xiaodong He\\
	JD AI Research, Beijing, China \\
	\texttt{\{sunhaipeng6, 
baojunwei, wuyouzheng1, hexiaodong\}@jd.com} \\
\\}

\begin{document}
\maketitle
\begin{abstract} 
Traditional end-to-end task-oriented dialog systems first convert dialog context into belief state and action state before generating the system response. The  system response performance is significantly affected by the quality of the belief state and action state. We first explore what dialog context representation is beneficial to improving the quality of the belief state and action state, which further enhances  the generated response quality. To tackle our exploration, we propose \textbf{Mars}, an end-to-end task-oriented dialog system with two contrastive learning strategies to model the relationship between dialog context and  belief/action state representations. Empirical results  show  dialog context representations, which are more different from semantic state representations, are more conducive to multi-turn task-oriented dialog.
Moreover, our proposed  Mars achieves state-of-the-art  performance on the MultiWOZ 2.0, CamRest676, and CrossWOZ\footnote{The code is available at \url{https://github.com/hpsun1109/Mars}.}.
\end{abstract}

%

\section{Introduction}
\label{sec:first}
Task-oriented dialog system~\cite{DBLP:journals/corr/abs-2003-07490} aims to assist users in completing  some specific tasks such as table reservations, hotel reservations,  ticket booking, and online shopping.
Traditional  task-oriented dialog system has been built through  dialog state tracking~\cite{lee-etal-2019-sumbt,wu-etal-2019-transferable}, dialog policy~\cite{DBLP:journals/corr/SchulmanWDRK17,takanobu-etal-2019-guided} and natural language generation~\cite{wen-etal-2015-semantically} tasks.
dialog state tracking  transfers dialog context to belief state, which is the structured semantic state capturing  the whole dialog context information. The belief state is used for the dialog system to query the database to  obtain  matched entities.
Dialog policy    selects an action state,  a semantic state guiding the dialog system to generate a system response based on the current dialog context and database information. System response is generated through a  natural language generation task.  
\begin{figure}[!t]
  \centering
  \includegraphics[width=3in]{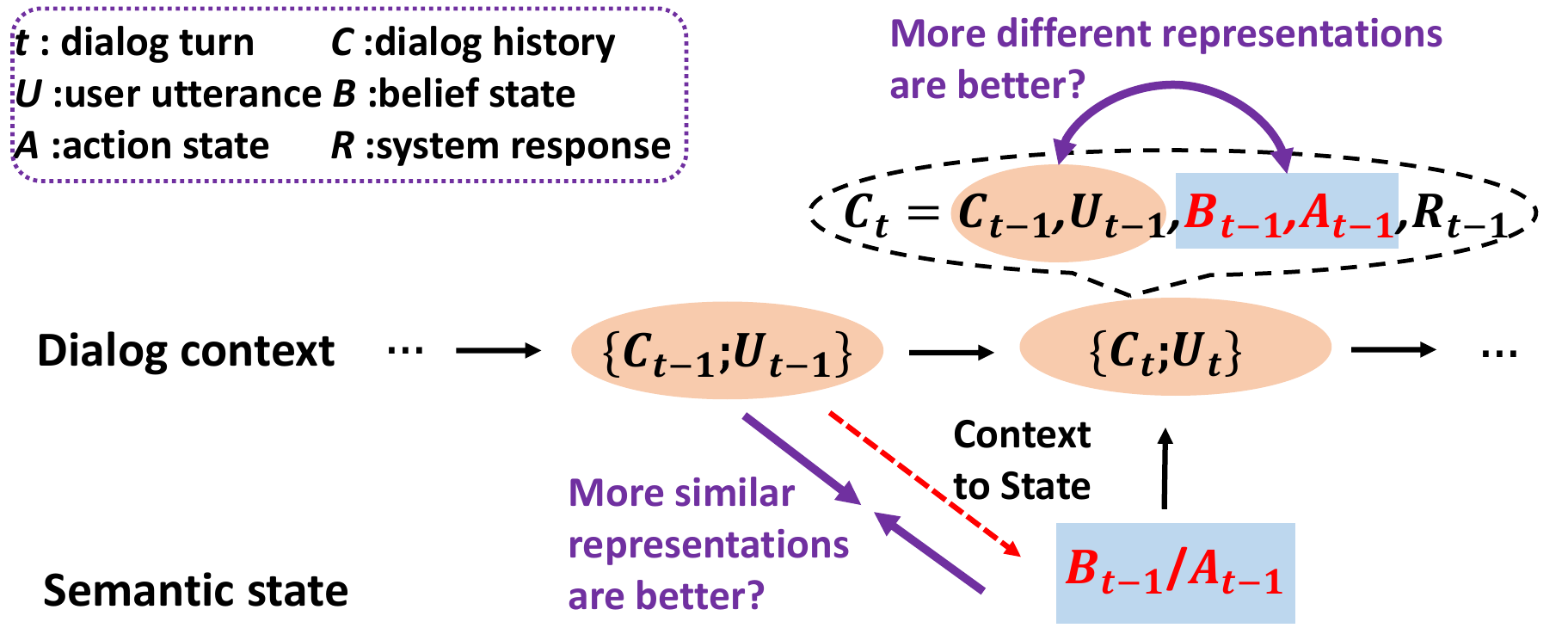}
  \caption{Illustration of the dialog context composition. Context to state represents dialog state tracking and dialog policy tasks. The previous dialog context $\{C_{t-1},U_{t-1}\}$ is included in the dialog context $\{C_{t},U_t\}$ of turn $t$. The database state is omitted for clarity.}
  \label{fig:representation}
\end{figure}

With the widespread application of large-scale pre-training models~\cite{devlin-etal-2019-bert,radford2019language,2020t5}, researchers gradually focus on the end-to-end task-oriented dialog system~\cite{lin-etal-2020-mintl,DBLP:conf/nips/Hosseini-AslMWY20,DBLP:conf/aaai/YangLQ21}, which  converts the whole dialog context into system response through multi-task training.
Generally, an end-to-end task-oriented dialog modeling task is formulated as a cascaded generation problem~\cite{DBLP:journals/corr/abs-2109-14739}. 
Before generating a system response, the end-to-end task-oriented dialog system must first  transfer dialog  context into belief and action states, respectively.  The quality of the belief state and action state greatly influence on the end-to-end task-oriented dialog performance\footnote{The detailed  analysis is provided in Appendix~\ref{sec:state}.}.

In this paper, we explore what dialog context representation is beneficial to improving the quality of the belief/action state, which further enhances the generated response quality.
As illustrated in Figure~\ref{fig:representation}, dialog context is recursively hybrid of previous dialog context and semantic states\footnote{These previous semantic states are  helpful references for the generation of the current turn~\cite{DBLP:conf/aaai/YangLQ21}.}, i.e., belief and action states, for multi-turn dialog.  
Intuitively, the representation of dialog context $\{C_{t-1},U_{t-1}\}$, which is more similar with that of semantic states $B_{t-1}/A_{t-1}$, is beneficial to generate semantic states of the  turn $t-1$. However, if their representations are too similar, there may be information redundancy in the representation of dialog context $\{C_{t},U_{t}\}$ in  turn $t$, as shown in Figure~\ref{fig:representation}. Thus we raise another conjecture: whether representations of dialog context, which are more different from that of semantic states, are more conducive to multi-turn task-oriented dialog?


To tackle our conjectures, we propose \textbf{Mars}, an end-to-end task-oriented dialog system with two contrastive learning strategies, i.e., pair-aware context\&state and group-aware context\&state contrastive learning, to model the relationship between dialog context and semantic states from two different levels.
Specifically, (1) the pair-aware context\&state contrastive learning strategy focuses more on narrowing the gap in the continuous representation space between dialog context and corresponding semantic states for the same dialog turn.
This strategy aims to obtain a continuous representation of the dialog context that is  semantically more consistent with that of its semantic states.
(2) Group-aware context\&state contrastive learning strategy enlarges the overall continuous representation margin between dialog context and semantic states.
The meaning behind this is to make representations between dialog context and semantic states more different.
Extensive experiments and analysis on the response generation and dialog state tracking tasks verify our raised conjectures and the effectiveness of Mars.
Mars achieves state-of-the-art performance on the MultiWOZ 2.0, CamRest676, and CrossWOZ.   Moreover, Mars  achieves remarkable performance in the low-resource scenario. Finally, we perform detailed error analysis and visualization to better apply our proposed Mars to real-world scenarios.

This paper primarily makes the following contributions:
 (1) We  explore what dialog context representation is beneficial to improving  task-oriented dialog performance.
	(2) We propose two contrastive learning strategies to model the relationship between dialog context and semantic state representations. 
	(3) Empirical results show   Mars achieves state-of-the-art  performance on the MultiWOZ 2.0, CamRest676, and CrossWOZ.

\section{Related Work}
End-to-end  task-oriented dialog systems~\cite{lei-etal-2018-sequicity,zhang-etal-2020-probabilistic,DBLP:conf/aaai/ZhangOY20} are established via copy-augmented seq2seq learning~\cite{gu-etal-2016-incorporating}.  \citet{DBLP:conf/aaai/ZhangOY20} proposes a multi-action data augmentation method to improve the diversity of generated system responses.
Large-scale pre-trained language models, including  BERT~\cite{devlin-etal-2019-bert},  GPT-2~\cite{radford2019language},   T5~\cite{2020t5}, and UniLM~\cite{DBLP:conf/nips/00040WWLWGZH19}, have been  demonstrated effective for improving the performance of task-oriented dialog systems~\cite{DBLP:conf/nips/Hosseini-AslMWY20,peng2020soloist,lin-etal-2020-mintl,DBLP:conf/aaai/YangLQ21,DBLP:journals/corr/abs-2103-06648,DBLP:journals/corr/abs-2111-14592} on MultiWOZ 2.0~\cite{budzianowski-etal-2018-multiwoz},  a large-scale English multi-domain task-oriented dialog dataset. Recently, auxiliary tasks and auxiliary dialog corpora have been introduced to further improve dialog modeling ability. MTTOD~\cite{lee-2021-improving-end} introduces a span prediction task to enhance the natural language understanding performance. BORT~\cite{2021BORTAnonymous} proposes reconstruction strategies to alleviate the error   propagation problem. 
PPTOD~\cite{DBLP:journals/corr/abs-2109-14739} proposes a dialog multi-task pre-training strategy to model task completion from  auxiliary heterogeneous dialog corpora. 
GALAXY~\cite{DBLP:journals/corr/abs-2111-14592} introduces a
dialog act prediction task to explicitly learn dialog policy from auxiliary dialog corpora.

Recently, contrastive Learning~\cite{DBLP:conf/cvpr/He0WXG20,DBLP:conf/icml/ChenK0H20,DBLP:conf/nips/GrillSATRBDPGAP20,DBLP:conf/cvpr/ChenH21} has attracted much attention in the computer vision community and has been applied to   natural language processing to enhance sentence representation learning~\cite{DBLP:journals/corr/abs-2005-12766,DBLP:journals/corr/abs-2012-15466,yan-etal-2021-consert,gao-etal-2021-simcse,giorgi-etal-2021-declutr}.  In contrast, we propose  contrastive learning strategies to model the relationship between dialog context and semantic state representations for task-oriented dialog. In addition, we don't introduce data augmentation methods, which are  used in most  contrastive learning works.

\section{Task-Oriented Dialog Framework}
\label{sec:third}

Generally, an end-to-end task-oriented dialog modeling task is formulated as a cascaded generation problem~\cite{DBLP:journals/corr/abs-2109-14739}. Before generating a system response, the end-to-end task-oriented dialog system would transfer dialog  context into belief state and action state, respectively.
Belief state is  a semantic state of dialog context, including dialog domain, slot name, and slot value. 
Action state is a  semantic state of system response, including dialog domain, dialog act, and slot name.  For example, the belief state is `\textit{[attraction] type theatre}', and the action state is `\textit{[attraction] [inform] name area}'.

\begin{figure}[t]
  \centering
  \includegraphics[width=2.85in]{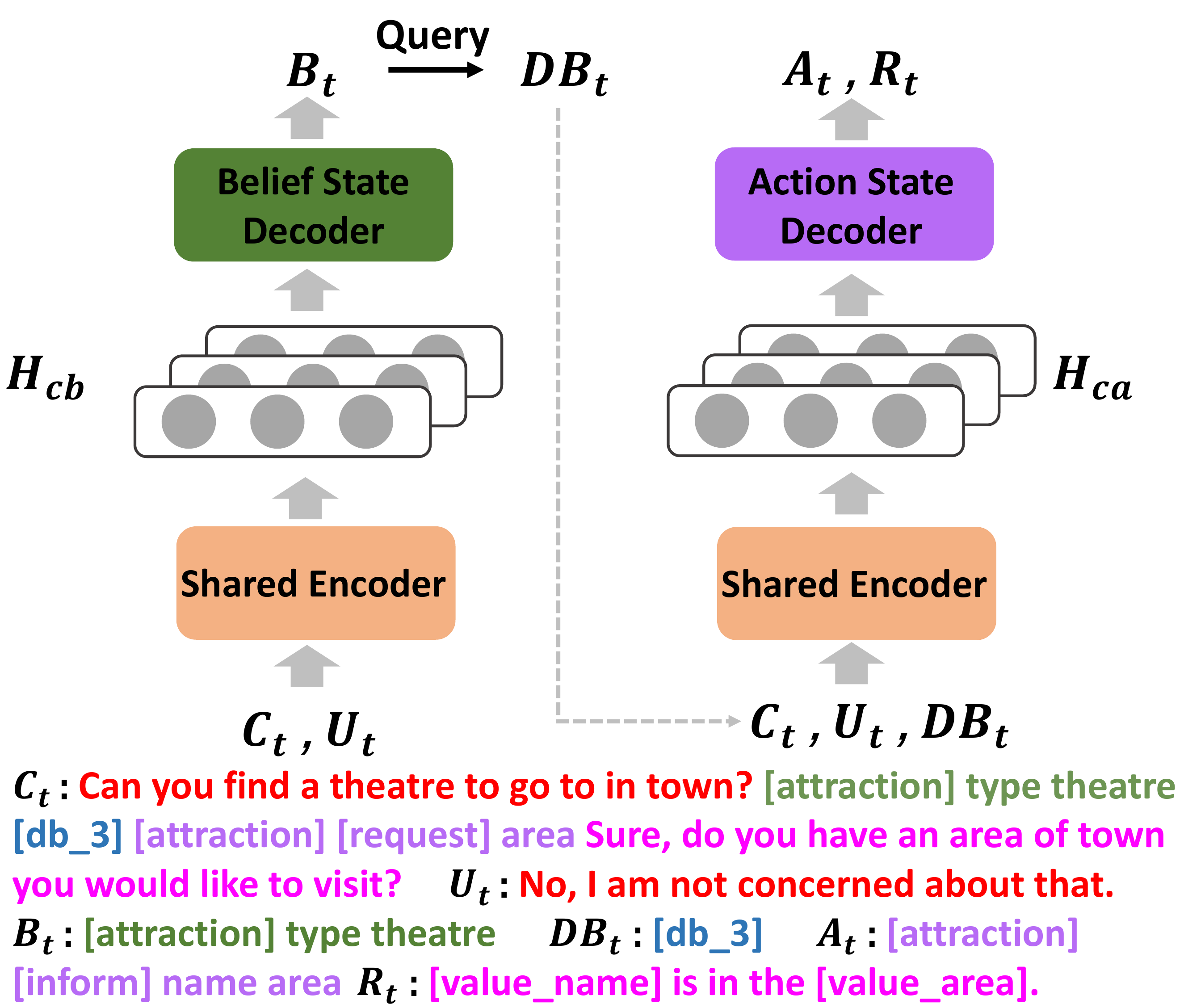}
  \caption{Illustration of a general  task-oriented dialog system. For clarity, we take dialog turn $t=1$ as an example. $C_t$ is formulated as $\{U_0,B_0,DB_0,A_0,R_0\}$. [db\_3] denotes the amount of matched entities.}
  \label{fig:baseline}
\end{figure}

We construct an end-to-end task-oriented dialog  system  via the seq2seq framework, including one shared encoder and two different decoders,
as illustrated in Figure \ref{fig:baseline}.
One shared encoder encodes dialog context, one decoder $decoder_b(\cdot)$ decodes belief state, and  another decoder $decoder_a(\cdot)$ decodes action state and system response.
Consider a dialog in turn $t$, dialog history $C_t$, which contains  dialog information for all previous turns, is formulated as $\{C_{t-1},U_{t-1},B_{t-1},DB_{t-1},A_{t-1},R_{t-1}\}$, where $U$ represents  the  user utterance, $B$ represents  the belief state, $DB$ represents  the database state, $A$ represents  the action state, and $R$ represents  the system response.

For end-to-end dialog modeling, a belief state is first generated. 
The dialog history $C_t$ and the  current user utterance $U_t$ are firstly encoded into hidden representation $H_{cb}$ through the shared encoder, and the belief state $B_t$ is generated through the belief state decoder:

{\footnotesize
\setlength{\abovedisplayskip}{0.005cm}
\setlength{\belowdisplayskip}{0.005cm}
\begin{equation}
\begin{aligned}
H_{cb} &= encoder(C_{t},U_t),\\
B_t &= decoder_b(H_{cb}).\\
\end{aligned}
\end{equation}}%
The dialog state tracking process is optimized by minimizing the following objective function:

{\footnotesize
\begin{equation}
\mathcal{L}_{B} =  -log P(B_t|C_{t},U_t).
\end{equation}}%
We use the generated belief state $B_t$ to query the specific database to achieve the database state $DB_t$, which means the amount of matched entities.

As described by MTTOD~\cite{lee-2021-improving-end}, the second decoder would be used to generate action state and system response simultaneously.
The combination of the dialog history $C_t$, the  current user utterance $U_t$, and  the database state $DB_t$ are encoded into hidden representation $H_{ca}$ through the shared encoder. The action state $A_t$  and system response $R_t$  are  generated in turn through the action state decoder:

{\footnotesize
\setlength{\abovedisplayskip}{0.005cm}
\setlength{\belowdisplayskip}{0.005cm}
\begin{equation}
\begin{aligned}
H_{ca} &= encoder(C_{t},U_t,DB_{t}),\\
A_t,R_t &= decoder_a(H_{ca}).\\
\end{aligned}
\end{equation}}%
 Therefore, the action state  and  response generation process is optimized by minimizing the following objective function:
 
 {\footnotesize
\setlength{\abovedisplayskip}{0.005cm}
\setlength{\belowdisplayskip}{0.005cm}
\begin{equation}
\mathcal{L}_{AR} =   -log P(A_t,R_t|C_{t},U_t,DB_{t}).
\end{equation}}%
 In summary, the entire end-to-end task-oriented dialog system can be optimized by minimizing:
 
 {\footnotesize
 \setlength{\abovedisplayskip}{0.005cm}
\setlength{\belowdisplayskip}{0.005cm}
\begin{equation}
\begin{aligned}
\mathcal{L}_{all} = \mathcal{L}_{B} + \mathcal{L}_{AR}.
\end{aligned}
\end{equation}}%
\section{Methodology}

\begin{figure*}[t]
  \centering
  \includegraphics[width=6.2in]{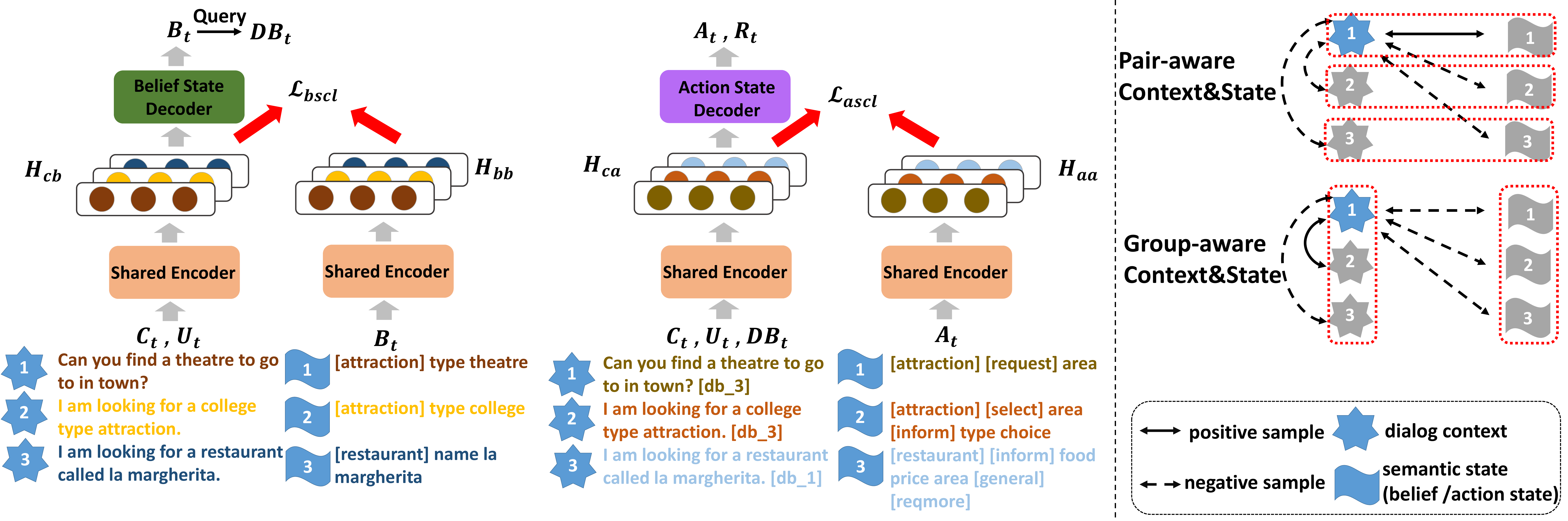}
  \caption{Illustration of the  task-oriented dialog system with contrastive learning strategies. We take dialog turn $t=0$ and batch size $N=3$ as an example.}  
  \label{fig:methods}
\end{figure*}

To tackle our conjectures and enhance the relationship modeling between dialog context and corresponding semantic state representations of   the task-oriented dialog system  described in Section~\ref{sec:third}, we propose two contrastive learning methods: pair-aware context\&state and group-aware context\&state  contrastive learning. Figure \ref{fig:methods} illustrates the architecture of a task-oriented dialog system with our  proposed  methods.
Generally, for any contrastive learning method,  contrastive learning objective functions $\mathcal{L}_{bscl}$ and $\mathcal{L}_{ascl}$ are added for dialog state tracking and response generation tasks, respectively, to enhancing the relationship modeling between dialog context and semantic state representations  during end-to-end dialog training. The general objective function can be reformulated as follows:

{\footnotesize
\setlength{\abovedisplayskip}{0.005cm}
\setlength{\belowdisplayskip}{0.005cm}
\begin{equation}
\begin{aligned}
\mathcal{L}_{all} &= \mathcal{L}_{B'} + \mathcal{L}_{AR'},\\
\mathcal{L}_{B'} &= \mathcal{L}_{B} +\lambda_{1}\mathcal{L}_{bscl} ,\\
\mathcal{L}_{AR'} &= \mathcal{L}_{AR} + \lambda_{2}\mathcal{L}_{ascl},\\
\end{aligned}
\end{equation}}%
where $\lambda_{1}$ and $\lambda_{2}$ are hyper-parameters that adjust the weight of the  objective functions.
\subsection{Pair-aware Context\&State Contrastive Learning}
To achieve dialog context representation, which is  semantically more consistent with  its semantic state representation,  we propose a pair-aware context\&state contrastive learning strategy (Mars-P) to close the continuous representation gap between  dialog context $\{C_t,U_t\}$ and corresponding semantic states, including belief state $B_t$  and action state $A_t$, for the same dialog turn.

We consider the dialog context $\{C_t,U_t\}$ and the belief state $B_t$ from the same dialog  to be as consistent as possible in the representation space, while the dialog context is as far away from other belief states as possible.
As illustrated in Figure~\ref{fig:methods}, the source continuous representation of the dialog context `\textit{can you find a theater to go to in town?}' should be similar to that of the belief state `\textit{[attraction] type theatre}' rather than other belief states  `\textit{[attraction] type college}' and `\textit{[restaurant] name la margherita}'.

Specifically, the belief state $B_t$ would be encoded into a hidden representation $H_{bb}$ through the shared encoder:

{\footnotesize
\setlength{\abovedisplayskip}{0.005cm}
\setlength{\belowdisplayskip}{0.005cm}
\begin{equation}
\begin{aligned}
H_{bb} &= encoder(B_{t}).\\
\end{aligned}
\end{equation}}%
For every dialog context input in a batch, we treat  the corresponding belief state from the same dialog  as a positive sample and other belief states and dialog contexts in the same batch as  negative samples.
Therefore, this dialog model is optimized by minimizing the objective function:

{\footnotesize
\setlength{\abovedisplayskip}{0.005cm}
\setlength{\belowdisplayskip}{0.005cm}
\begin{equation}
\begin{aligned}
\mathcal{L}_{bscl} &\triangleq \mathcal{L}_{bscl\_P}\\&= -log \frac{e^{cos(H_{cb}^i, H_{bb}^i)/T}}{\sum\limits_{\substack{k=1\\k\neq i}}^{N}\! e^{cos(H_{cb}^i,H_{cb}^k)/T} \!+ \!\sum\limits_{k=1}^{N}\! e^{cos(H_{cb}^i,H_{bb}^k)/T}},
\end{aligned}
\end{equation}}%
where $cos(\cdot)$ denotes the cosine similarity function. $T$ is a temperature hyperparameter. $N$ is the batch size.
In a batch, $H_{cb}^i$ denotes the $i$th dialog context hidden representation after average pooling, and $H_{bb}^k$ denotes the $k$th  belief state hidden representation after average pooling.

During response generation,  
we would close the continuous representation gap of dialog context $\{C_t,U_t,DB_t\}$ and action state $A_t$. 
As illustrated in Figure~\ref{fig:methods}, the source continuous representation of the user utterance `\textit{i am looking for a restaurant called la margherita.}' 
and database information `\textit{[$db_1$]}' should be similar to that of the action state `\textit{[restaurant] [inform] food price area [general] [reqmore]}' rather than other action states  `\textit{[attraction] [request] area}' and `\textit{[attraction] [select] area [inform] type choice}'.
Specifically, the action state $A_t$ would be encoded into a hidden representation $H_{aa}$ through the shared encoder:

{\footnotesize
\setlength{\abovedisplayskip}{0.005cm}
\setlength{\belowdisplayskip}{0.005cm}
\begin{equation}
\begin{aligned}
H_{aa} &= encoder(A_{t}).\\
\end{aligned}
\end{equation}}%
For every dialog context input  in a batch, we treat the corresponding action state from the same dialog  as  a positive sample and other action states and dialog contexts in the same batch as  negative samples.
Therefore, this dialog model is optimized by minimizing the objective function:

{\footnotesize
\setlength{\abovedisplayskip}{0.005cm}
\setlength{\belowdisplayskip}{0.005cm}
\begin{equation}
\begin{aligned}
\mathcal{L}_{ascl} &\triangleq \mathcal{L}_{ascl\_P}\\&= -log \frac{e^{cos(H_{ca}^i, H_{aa}^i)/T}}{\sum\limits_{\substack{k=1\\k\neq i}}^{N}\! e^{cos(H_{ca}^i,H_{ca}^k)/T} \!+ \!\sum\limits_{k=1}^{N}\! e^{cos(H_{ca}^i,H_{aa}^k)/T}},
\end{aligned}
\end{equation}}%
where $H_{ca}^i$ denotes the $i$th dialog context hidden representation after average pooling, and $H_{aa}^k$  denotes the $k$th  action state hidden representation after average pooling.

\subsection{Group-aware Context\&State  Contrastive Learning}
To explore whether representations of dialog context, which are more different from that of semantic states, are more conducive to multi-turn task-oriented dialog, 
 we propose a group-aware context\&state contrastive learning strategy (Mars-G). 
Takes turn $t$ as an example,
Mars-G enlarges the overall continuous representation margin between dialog context and semantic states, regardless of the pairing relationship between specific dialog context, e.g., $\{C_i,U_i\}$, and semantic states, e.g.$B_i/A_i$ (turn $i=0,...,t$).
The meaning behind is to make representations between dialog context and semantic states more different, which makes it easy to distinguish dialog context $\{C_i,U_i\}$ and the corresponding semantic states $B_i/A_i$ (turn $i=0,...,t$) inside the entire dialog context $\{C_{t+1},U_{t+1}\}$ and achieve much richer dialog context representations.

Specifically, for every dialog context input, we treat all semantic states in the same batch as negative samples and any one dialog context in the same batch as a positive sample.
Besides, considering that every dialog input contains a unique context, narrowing the in-batch context distance makes it hard to distinguish different contexts, which may be counterproductive to deriving  the context representation. To resolve such an issue, we also select the rest in-batch dialog context inputs except the positive one as negative samples for every dialog context input.
Therefore,  the contrastive learning objective function can be reformulated as:

{\footnotesize
\setlength{\abovedisplayskip}{0.005cm}
\setlength{\belowdisplayskip}{0.005cm}
\begin{equation}
\begin{aligned}
\mathcal{L}_{bscl} &\triangleq \mathcal{L}_{bscl\_G}\\&= -log \frac{e^{cos(H_{cb}^i, H_{cb}^j)/T}}{\sum\limits_{\substack{k=1\\k\neq i}}^{N}\! e^{cos(H_{cb}^i,H_{cb}^k)/T} \!+ \!\sum\limits_{k=1}^{N}\! e^{cos(H_{cb}^i,H_{bb}^k)/T}},
\end{aligned}
\end{equation}%
\begin{equation}
\begin{aligned}
\mathcal{L}_{ascl} &\triangleq \mathcal{L}_{ascl\_G}\\&= -log \frac{e^{cos(H_{ca}^i, H_{ca}^j)/T}}{\sum\limits_{\substack{k=1\\k\neq i}}^{N}\! e^{cos(H_{ca}^i,H_{ca}^k)/T} \!+ \!\sum\limits_{k=1}^{N}\! e^{cos(H_{ca}^i,H_{aa}^k)/T}},
\end{aligned}
\end{equation}}%
where $H_{cb}^j$ and $H_{ca}^j$ denote the $j$th ($j \neq i$) dialog context hidden representations after average pooling.

\begin{table*}[ht]
  \centering
  \scalebox{.65}{
	\begin{tabular}{lcccccccc}
		\toprule
		\multirow{2}{*}{\bf Model} &\multirow{2}{*}{\bf Pre-trained} &\multirow{2}{*}{\bf Extra corpora}&\multicolumn{1}{c}{\bf  Dialog state tracking}&\multicolumn{5}{c}{\bf Response Generation}\\  
		
	&  &	& \bf Joint Accuracy & \bf Act F1& \bf Inform & \bf Success & \bf BLEU & \bf Combined\\ 
		\midrule
		DAMD~\cite{DBLP:conf/aaai/ZhangOY20}&-&no &-&-&57.9 &47.6 & 16.4 &69.2\\
				LABES~\cite{zhang-etal-2020-probabilistic}&-&no&-&-&68.5&58.1&18.9&82.2\\
				AuGPT~\cite{DBLP:journals/corr/abs-2102-05126}&GPT-2&yes&-&-&76.6&60.5&16.8&85.4\\
		MinTL~\cite{lin-etal-2020-mintl}& T5-small&no&51.2&-&73.7&65.4&19.4 &89.0\\
		
		SOLOIST~\cite{peng2020soloist}&GPT-2&yes&53.2&-& 82.3&72.4&13.6&91.0\\
		
		DoTS~\cite{DBLP:journals/corr/abs-2103-06648}&BERT-base&no&-&-&80.4&68.7&16.8&91.4\\
		UBAR~\cite{DBLP:conf/aaai/YangLQ21}  &DistilGPT2&no&52.6 &-&  83.4&70.3&17.6 & 94.5\\
     PPTOD~\cite{DBLP:journals/corr/abs-2109-14739}&T5-base&yes&53.4&-&83.1&72.7&18.2&96.1\\
		BORT~\cite{2021BORTAnonymous}& T5-small&no&54.0&-& 85.5 &   77.4 & 17.9 &  99.4 \\
			MTTOD~\cite{lee-2021-improving-end}&T5-base&no&53.6&-& 85.9&76.5&19.0&100.2   \\
						GALAXY~\cite{DBLP:journals/corr/abs-2111-14592}&UniLM-base &yes &-&-&85.4&75.7& 19.6 & 100.2\\ 
		   \cdashline{1-9}[1pt/2pt]
		    Baseline& T5-small&no&53.8&53.0&      83.2&70.3&19.4&96.2\\   
		    Mars-P& T5-small&no&  54.4 & \bf 53.9  &  86.6&  75.5& 19.6 &100.7\\

		   Mars-G& T5-small&no&\bf 55.1 & 53.7&\bf88.9& \bf 78.0&\bf 19.9& \bf 103.4\\

		\bottomrule
	\end{tabular}}\caption{Comparison of end-to-end models evaluated on MultiWOZ 2.0. The results of previous work are reported on the official leaderboard of MultiWOZ (\url{https://github.com/budzianowski/multiwoz}).  \label{tab:main_results}}
\end{table*}
\section{Experiments}  
\label{sec:experiments}
\subsection{Datasets and Evaluation Metrics}
\label{sec:evaluation}
We conduct experiments on three task-oriented dialog datasets: MultiWOZ 2.0~\cite{budzianowski-etal-2018-multiwoz},  CamRest676~\cite{wen-etal-2017-network}, and CrossWOZ~\cite{zhu-etal-2020-crosswoz}. 
MultiWOZ 2.0~\cite{budzianowski-etal-2018-multiwoz} and CamRest676~\cite{wen-etal-2017-network} are English task-oriented dialog datasets. 
CrossWOZ~\cite{zhu-etal-2020-crosswoz} is a Chinese multi-domain task-oriented dialog dataset.
A detailed description  of the datasets is provided in Appendix~\ref{sec:dataset}.

We test our proposed Mars on two benchmark task-oriented dialog tasks: end-to-end dialog modeling response generation  and dialog state tracking. We evaluate the performance of response generation on MultiWOZ 2.0 and CamRest676.
Inconsistencies exist between previous task-oriented dialog works  in data preprocessing and evaluation metrics on  MultiWOZ 2.0~\cite{nekvinda-dusek-2021-shades}.
To fairly compare  our experiments with previous work,  we use the pre-processing strategy~\cite{DBLP:conf/aaai/ZhangOY20} and  the standalone standardized evaluation script released by ~\citet{nekvinda-dusek-2021-shades}. We follow the automatic evaluation metrics to evaluate the response quality for  task-oriented dialog system on MultiWOZ 2.0.
\textbf{Inform  rate} measures whether  a dialog system has provided an accurate entity; \textbf{Success rate} measures whether a dialog system has provided an accurate entity and answered all requested information; \textbf{BLEU score}~\cite{papineni-etal-2002-bleu}, which is computed with references, which have been obtained from the delexicalized MultiWOZ 2.2 span annotations, measures  the fluency of the generated response; \textbf{Combined score}, which is calculated  by $ (Inform + Success) \times 0.5 + BLEU$, measures the overall quality of the dialog system. Moreover, we use the \textbf{Act F1} to measure the accuracy of generated action states.
To make our experiments comparable with previous work \cite{zhang-etal-2020-probabilistic,DBLP:journals/corr/abs-2111-14592} on CamRest676,
we use the same pre-processing strategy and use \textbf{Inform  rate}, \textbf{Success F1}, \textbf{BLEU score}, and \textbf{Combined score}, which is computed by $ (Inform + Success F1) \times 0.5 + BLEU$, to evaluate the response quality for the task-oriented dialog system. The  success rate
whether if the system answered all requested
information to assess  recall, while Success F1 balances recall and precision.

We evaluate the performance of dialog state tracking on MultiWOZ 2.0 and CrossWOZ. We use the \textbf{joint goal accuracy} to measure the accuracy of generated belief states.

\subsection{Settings}
We use a pre-trained T5 language model~\cite{2020t5} to initialize the dialog system based on the HuggingFace Transformers library~\cite{wolf-etal-2020-transformers} and follow the settings of ~\citet{lee-2021-improving-end}. 
We select   T5-small~\cite{2020t5} for MultiWOZ 2.0 and CamRest676 and T5-base-Chinese~\cite{2020t5,zhao-etal-2019-uer} for CrossWOZ.
The batch size is  8. 
The AdamW optimizer~\cite{DBLP:conf/iclr/LoshchilovH19} optimizes  the model parameters with linear learning rate decay. The initial learning rate is 0.0005, and the ratio of warm up is 0.2.
The hyper-parameters $\lambda_{1}$ and $\lambda_{2}$ are set to 1 and 0.1, respectively. $T$ is set to 0.1 for Mars-P, and $T$ is set to 0.5 for Mars-G. The hyper-parameter  analysis is provided in Appendix~\ref{sec:parameter}.
We train all dialog systems on one NVIDIA A100 GPU  for 10 epochs and select the checkpoint model with the best performance on the validation dataset.  One model is trained for approximately five hours. In addition, the model is trained for 20 epochs for the  low resource scenarios.
  The description  of baseline systems is provided in Appendix~\ref{sec:baseline}. Another baseline is the general architecture of a task-oriented dialog system, as illustrated in Figure~\ref{fig:baseline}.

\subsection{Main Results}

\begin{figure}[!t]
	\centering
	\subfigure{  \begin{minipage}[b]{0.3\linewidth}
\includegraphics[width=\columnwidth]{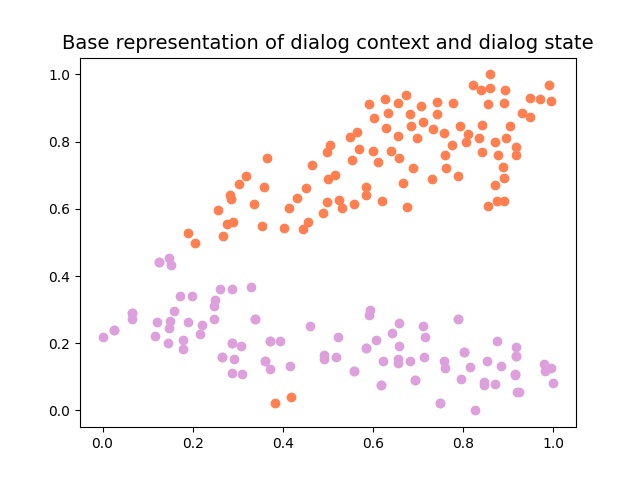}
		\end{minipage}
		
	}
	\subfigure{
		\begin{minipage}[b]{0.3\linewidth}
\includegraphics[width=\columnwidth]{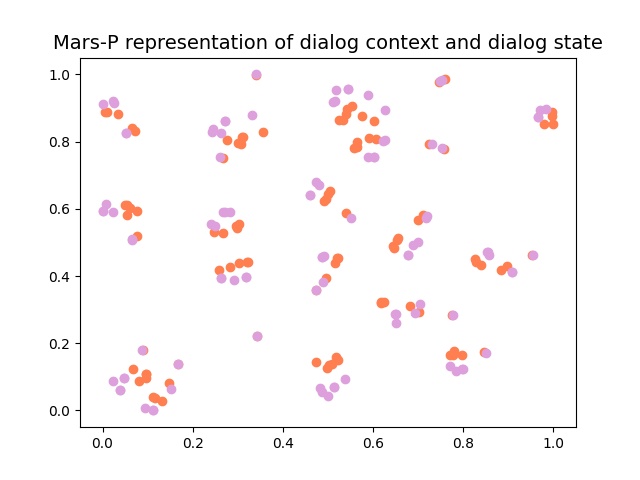}
		\end{minipage}
	}
	\subfigure{
		\begin{minipage}[b]{0.3\linewidth}
\includegraphics[width=\columnwidth]{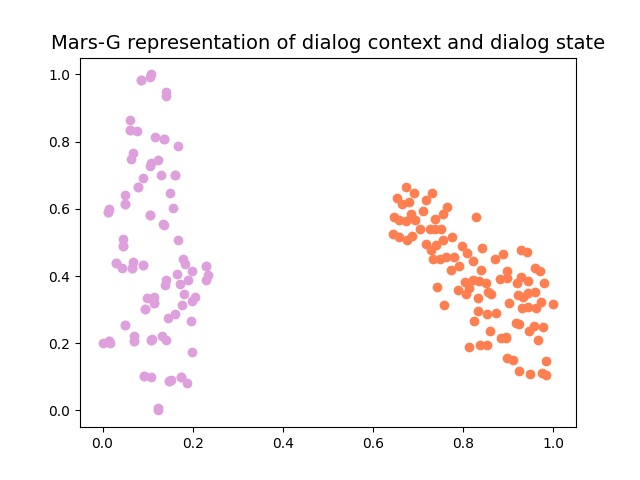}
		\end{minipage}
	}
		\subfigure{
		\begin{minipage}[b]{0.3\linewidth}
\includegraphics[width=\columnwidth]{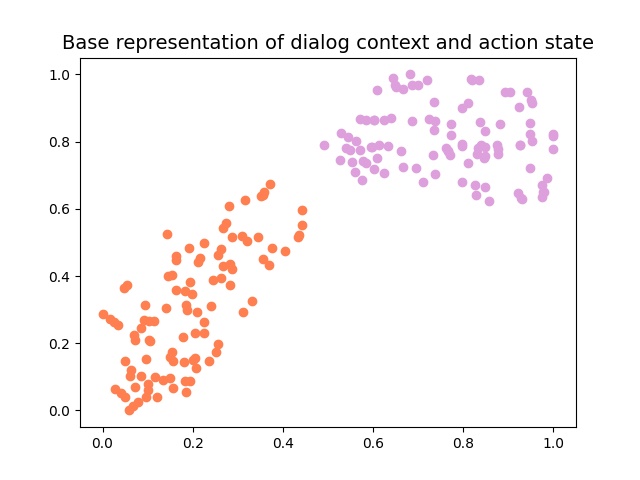}
		\end{minipage}
	}
	\subfigure{
		\begin{minipage}[b]{0.3\linewidth}
\includegraphics[width=\columnwidth]{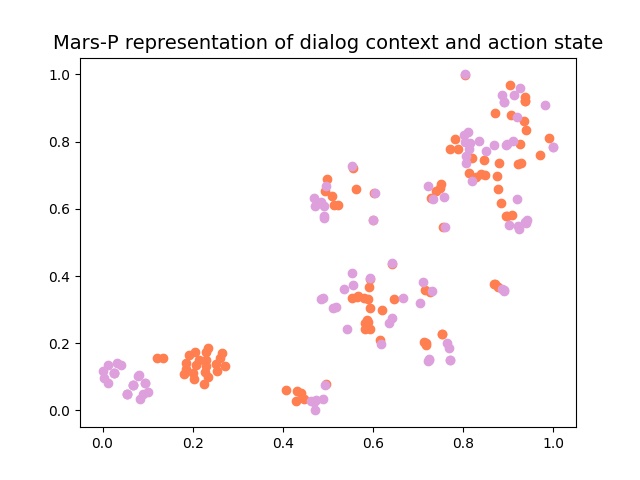}
		\end{minipage}
	}
	\subfigure{\begin{minipage}[b]{0.3\linewidth}
\includegraphics[width=\columnwidth]{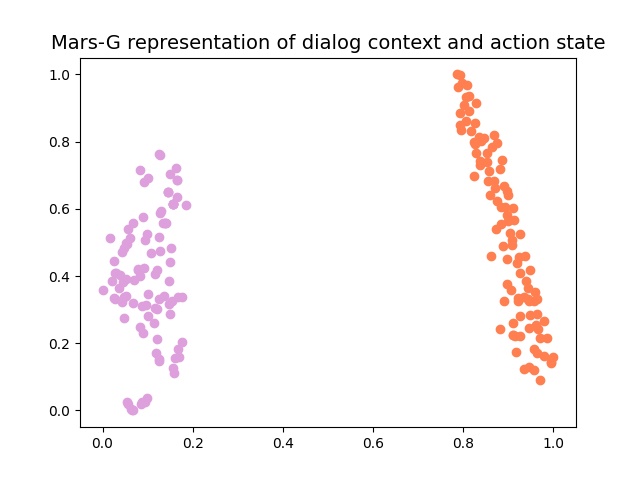}
		\end{minipage}
	}
\caption{Visualization of dialog context and semantic state representations using t-sne. The three sub-figures on the first row show baseline/Mars-P/Mars-G representations of dialog context and dialog state. The three sub-figures on the second row show baseline/Mars-P/Mars-G representations of dialog context and action state. The coral dots denote dialog context representations. The plum dots denote semantic state representations.  We plot 100 dialog examples.}\label{tsne}
\end{figure}

The detailed inform rates, success rates, BLEU scores, combined scores, act F1 scores, and joint goal accuracies for end-to-end task-oriented dialog models on the MultiWOZ 2.0 benchmark are presented in Table~\ref{tab:main_results}. 
Our re-implemented baseline system performs comparable with PPTOD~\cite{DBLP:journals/corr/abs-2109-14739}, 
and our proposed Mars-P and Mars-G outperform our re-implemented baseline system  by 4.5 and 7.2 combined scores.
Moreover,  Mars-G, which doesn't use auxiliary corpora, substantially outperforms the previous state-of-the-art GALAXY~\cite{DBLP:journals/corr/abs-2111-14592} and 	MTTOD~\cite{lee-2021-improving-end} by 3.2 combined scores, achieving the state-of-the-art performance in terms of inform rate, success rate, BLEU score, and combined score.  In addition, Mar-G achieves the highest joint goal accuracy among the end-to-end task-oriented dialog systems, outperforming BORT~\cite{2021BORTAnonymous} by 1.1 points.  Compared with the baseline system, Mars-P and  Mars-G achieve a better act F1 score.
This demonstrates  our proposed contrastive learning could effectively improve the quality of the belief state and action state, which further improves  the generated response quality. 
Regarding the two proposed methods, Mars-G performs  better than Mars-P. 
Figure~\ref{tsne} displays the visualization of dialog context and semantic state representations using t-sne.  Compared with the baseline system, Mars-P could achieve dialog context representation that is  semantically more consistent with  its semantic state representation while
Mars-G could make representations between dialog context and semantic states more different.
These verify dialog context representations, which are more different from  semantic state representations, are more beneficial to achieving  task completion of task-oriented dialog.
Further dialog context representation analysis  is provided in   Appendix~\ref{sec:Further}.

\begin{table}[!t]
  \centering
  \scalebox{.6}{
	\begin{tabular}{lcccc}
		\toprule
		\bf Model  & \bf Match & \bf Success F1& \bf BLEU & \bf Combined\\ 
		\midrule
Sequicity~\cite{lei-etal-2018-sequicity}& 92.7  &85.4& 25.3 &114.4 \\
		LABES~\cite{zhang-etal-2020-probabilistic} &   96.4  &  82.3 &  25.6 &  115.0 \\
		SOLOIST~\cite{peng2020soloist} &94.7 & 87.1 &25.5& 116.4 \\
 GALAXY~\cite{DBLP:journals/corr/abs-2111-14592}&\bf 98.5&87.7&24.2&	117.3\\
        \cdashline{1-5}[1pt/2pt]

        Mars-P & 97.0 & 87.2 &  25.9 & 118.0\\
         Mars-G& 96.2&\bf 89.6&  \bf 26.1& \bf 119.0\\

		\bottomrule
	\end{tabular}}\caption{Comparison of end-to-end  task-oriented dialog systems on CamRest676.\label{tab:camrest}}
\end{table}

Table~\ref{tab:camrest} presents the performance of task-oriented dialog systems on the CamRest676. Mars-G outperforms the previous state-of-the-art GALAXY~\cite{DBLP:journals/corr/abs-2111-14592} by 1.7 combined scores, achieving the state-of-the-art performance in terms of success F1, BLEU score, and combined score. 

\begin{table}[!t]
  \centering
  \scalebox{.62}{
	\begin{tabular}{lc}
		\toprule
		\bf Model & \bf Joint Accuracy \\ 
		\midrule
		TRADE~\cite{wu-etal-2019-transferable}&36.1\\
		BART-CSP~\cite{DBLP:journals/corr/abs-2111-02574}&53.6\\
GEEX~\cite{li-etal-2021-generation}&54.7\\
		\cdashline{1-2}[1pt/2pt]
		Mars-P&  59.3  \\
Mars-G& \bf 59.8 \\
		\bottomrule
	\end{tabular}}\caption{Comparison of dialog state tracking performance on CrossWOZ.\label{tab:dst}}
\end{table}

Table~\ref{tab:dst} reports the dialog state tracking performance  on CrossWOZ.   Mars-P and Mars-G substantially outperform the previous state-of-the-art GEEX~\cite{li-etal-2021-generation}  by  4.6 and 5.1 points, achieving 59.3 and 59.8 joint goal accuracy. This further indicates that our proposed  contrastive learning strategies could improve belief state learning ability, and Mars has good generalization ability. In addition, we provide an example to visualize  our proposed Mars-G's dialog state tracking process in  Appendix~\ref{app:vis}.


\subsection{Ablation Study}
Table~\ref{tab:ablation} shows the performance of the different components of  Mars-P and Mars-G. Both state  modules of Mars-P and Mars-G could improve the performance of the  dialog system.
Regarding  two modules of contrastive learning strategies Mars-P and Mars-G, the action state  module performs  better than the belief state  module by 1.7 and  1.6 combined scores, respectively, because the quality of the  action state has a  more direct impact on the response generation quality and action state  module could improve action state learning ability. 
Moreover, the combination of both modules  can complement each other to further improve end-to-end dialog modeling performance. The further ablation analysis  is provided in   Appendix~\ref{app:sim}.

\begin{table}[!t]
  \centering
  \scalebox{.59}{
	\begin{tabular}{lcccc}
		\toprule
		\bf Model & \bf Inform & \bf Success & \bf BLEU & \bf Combined\\ 
		\midrule
Mars-G&88.9 & 78.0 & 19.9 & 103.4\\
\;\;\;\;\;\;\;\;\;\;w/o BSC &88.3&76.6&19.5& 102.0 \\
\;\;\;\;\;\;\;\;\;\;w/o ASC&86.3&75.1&19.7&100.4\\

\midrule
Mars-P&86.6&75.5&19.6&100.7\\
\;\;\;\;\;\;\;\;\;\;w/o BSC & 85.4 & 75.0& 19.7 & 99.9 \\
\;\;\;\;\;\;\;\;\;\;w/o ASC&83.7& 73.0& 19.8&98.2\\

\midrule
Baseline& 83.2&70.3&19.4&96.2\\

		\bottomrule
	\end{tabular}}\caption{The performance of the different components of our proposed methods on MultiWOZ 2.0.  BSC represents  the belief state  module of contrastive learning,  and ASC represents  the action state module.\label{tab:ablation}}
\end{table}


\begin{table*}[ht]
   \small
   \centering
   \scalebox{.625}{
	\begin{tabular}{lcccccccccccccccc}
		\toprule
		\multirow{2}*{\bf Model} &\multicolumn{4}{c}{\bf 5\% } &\multicolumn{4}{c}{\bf 10\% } &   \multicolumn{4}{c}{\bf 20\%}&  \multicolumn{4}{c}{\bf 50\% }\\ 
\cmidrule(lr){2-5} \cmidrule(lr){6-9} \cmidrule(lr){10-13} \cmidrule(lr){14-17} 
		& \bf Inform  & \bf Success & \bf BLEU & \bf Combined& \bf Inform  & \bf Success & \bf BLEU& \bf Combined& \bf Inform  & \bf Success & \bf BLEU& \bf Combined& \bf Inform  & \bf Success & \bf BLEU& \bf Combined\\ 
		\midrule

DAMD &   36.8&   17.3&  11.2&   38.3& 40.9  &23.0   & 12.2  &    44.2&  48.3 & 30.3 &   14.2&  53.5&  58.8    & 44.3  &15.7    &67.3   \\
MinTL &  52.5 & 38.1 & 13.9 &59.2  &55.5  & 44.9  & 15.6 &65.8  &  64.3 & 54.9  &  16.2& 75.8 & 70.3&  62.2 & 18.0  &  84.3  \\
UBAR & 37.4 & 22.1 & 11.3  & 41.1  &  50.3& 34.2&  13.5 &55.8 & 65.5  & 48.7  &14.5   &71.6   &77.6   & 63.3  & 16.3  & 86.8 \\  
MTTOD   & 54.3&37.4& 11.3  & 57.2&   66.9&  55.2 & 13.8  &  74.9  & 75.0&  \bf 63.3& 14.3    &  83.5 & 78.5 &  67.5 &  15.2  &  88.2       \\
PPTOD &    \bf 65.5&    \bf  48.3    & \bf 14.3&   \bf 71.2       &   68.3 &  53.7&  \bf 15.7  &76.7& 72.7& 59.2&16.3&82.3&  74.8 &  62.4& 17.0&85.6 \\
\cdashline{1-17}[1pt/2pt]
Mars-G    &  57.6  & 43.4 & 13.9 & 64.4& \bf 69.4 & \bf 55.3  & 15.6 &  \bf 78.0 & \bf 76.7  &  62.9 &  \bf 17.2 & \bf  87.0&  \bf  82.2& \bf 71.2 &  \bf 18.6 & \bf 95.3     \\

		\bottomrule
	\end{tabular}}\caption{Comparison of task-oriented dialog systems  in the low resource scenarios on  MultiWOZ 2.0. 5\% (400 dialogs), 10\% (800 dialogs), 20\% (1600 dialogs),   50\% (4000 dialogs) of training data is used to train each model.\label{tab:low_resource}}

\end{table*}
\subsection{ Dialog Turn Analysis}

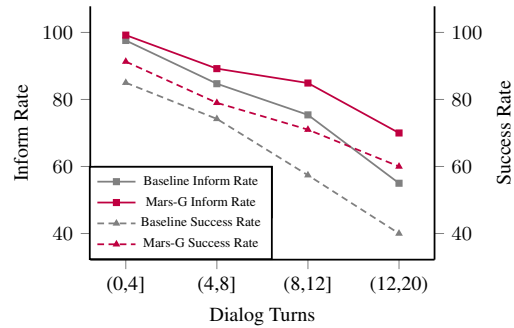
\begin{figure}[!t]

	\setlength{\abovecaptionskip}{0pt}
	 	\begin{center}
	 	  \scalebox{.95}{
			\pgfplotsset{height=5.6cm,width=8.5cm,compat=1.14,every axis/.append style={thick},every axis legend/.append style={
at={(0.53,0.37)}},legend columns=1}
		\scalebox{.88}{	\begin{tikzpicture}
			\tikzset{every node}=[font=\small]
			\begin{axis}
			[width=7cm,enlargelimits=0.13, tick align=outside, xticklabels={ {(0,4]},{(4,8]},{(8,12]},{(12,20)}},
 axis y line*=left,
xtick={0,1,2,3},
 ylabel={Inform Rate},
 axis x line*=left, 
 ylabel style={align=left},xlabel={ Dialog Turns},ymax=100,ymin=40,font=\small]
			\addplot+ [sharp plot,mark=square*,mark size=1.2pt,mark options={solid,mark color=gray}, color=gray] coordinates 
			{ (0,97.6)(1,84.7)(2,75.4)(3,55.0) };\label{plot_bb}
			\addlegendentry{\tiny Baseline Inform Rate} 
						\addplot+ [sharp plot,mark=square*,mark size=1.2pt,mark options={solid,mark color=purple}, color=purple] coordinates
			{ (0,99.2)(1,89.2)(2,84.9)(3,70.0) };\label{plot_aa}
			\addlegendentry{\tiny Mars-G Inform Rate}

\end{axis}
\begin{axis}
[width=7cm,enlargelimits=0.13, tick align=outside,  xticklabels={ {(0,4]},{(4,8]},{(8,12]},{(12,20)}},
axis y line*=right,
axis x line=none,
xtick={0,1,2,3},  
 ylabel={Success Rate},xlabel={ Dialog Turns},ymax=100,ymin=40,font=\small]
 
\addlegendimage{/pgfplots/refstyle=plot_bb}\addlegendentry{\tiny Baseline Inform Rate} 
\addlegendimage{/pgfplots/refstyle=plot_aa}\addlegendentry{\tiny Mars-G Inform Rate}
		\addplot+ [sharp plot,densely dashed,mark=triangle*,mark size=1.2pt,mark options={solid,mark color=gray}, color=gray] coordinates
			{ (0,85.0 )(1,74.2)(2,57.4)(3,40.0) };
			\addlegendentry{\tiny Baseline Success Rate}
						\addplot+ [sharp plot,densely dashed,mark=triangle*,mark size=1.2pt,mark options={solid,mark color=purple}, color=purple] coordinates
			{ (0,91.3)(1,79.0)(2,71.0)(3,60.0) };
			\addlegendentry{\tiny Mars-G Success Rate}
			
			\end{axis}
			\end{tikzpicture}}}

		\end{center}
  		\caption{Performance of dialog systems on the test set with respect to  different dialog turns.  }\label{fig:turn}
\end{figure} 

To better assess the effectiveness of  our proposed  contrastive learning strategies, we investigate the performance (inform rate and success rate) of Mars-G and the baseline system on the test set with respect to different dialog turns.  Specifically, we divide
each test set into four groups according to the dialog turn. As shown in Figure~\ref{fig:turn}, Mars-G is superior to the baseline system in every dialog turn group. This indicates our proposed contrastive learning strategies are  beneficial to task-oriented dialog modeling. Especially, as dialog turn increases, the performance of the baseline system
decreases rapidly, and the performance gap between the baseline system and our proposed Mars-G is increased. Because the baseline system  is hard to model long-range semantic dependencies to generate inaccurate semantic states and system responses. In contrast,  Mars-G enhances the relationship modeling between dialog context and  semantic state representations  and achieves better dialog context representations to capture long-range semantic dependencies in the long dialog turns.

\subsection{Low Resource Scenario Analysis}

To investigate the performance of task-oriented dialog systems in the  low resource scenario, we choose 5\%, 10\%, 20\%, and 50\% of training dialog sessions   to do stimulated experiments on the MultiWOZ 2.0. 
Considering the inconsistency of data distribution with different random seeds in the  stimulated low resource scenario, we  re-implement all baseline systems with the same random seed to ensure the consistency of data distribution.
In addition, we train all dialog systems five times with different random seeds and report the average scores in Table~\ref{tab:low_resource}. The detailed results of five runs are provided in Appendix~\ref{app:low}.
As shown in Table~\ref{tab:low_resource},  PPTOD achieves the best performance in the extreme low-resource  scenario (5\% training data) because  auxiliary corpora  used in PPTOD have many similar dialog sessions with MultiWOZ 2.0 and this benefits PPTOD in the stimulated low-resource  scenario. In contrast, Mars-G doesn’t use auxiliary corpora to improve performance in the low-resource  scenario.
Apart from this,  Mars-G substantially outperforms all baseline systems  in other low-resource scenarios.
Moreover, Mars-G trained on the 50\% training data performs better than some baseline systems such as MinTL and UBAR trained on all training data, as shown in Table~\ref{tab:main_results}. These further demonstrate that Mars-G is robust, achieving comparable performance in the low-resource scenario.

\subsection{ Error Analysis}

To better apply our proposed Mars-G to real-world scenarios, we perform error analysis based on inform rate (informable slot) and success rate (requestable slot). In detail,  we randomly extract  40 inaccurate dialog sessions from the MultiWOZ 2.0 testing set, respectively. 
\begin{figure}[t!]
  \centering \scalebox{.88}{
  \subfigure{
  \begin{minipage}{.4\linewidth}
  \centering
  \scalebox{.46}{
\begin{tikzpicture}
\begin{axis}
[
    xbar, enlarge y limits=0.1,
enlarge x limits={0.1,upper},
    xlabel={\ Dialog Sessions}, symbolic y coords={  Attraction,Restaurant,Hotel,Train,Taxi},
    ytick=data, nodes near coords, 
    nodes near coords align={horizontal},axis y line*=left,
		axis x line*=left, yticklabel style={rotate=60}]
\addplot  +[color=teal] coordinates { (8,Train) (13,Restaurant) (13,Hotel) (10,Attraction) (0,Taxi)};
\addplot +[color=gray] coordinates {(22,Train) (19,Restaurant) (19,Hotel) (16,Attraction) (6,Taxi)};
\end{axis}
\end{tikzpicture}}
\centerline{(a)}
  \end{minipage}
    }
  \subfigure{
\begin{minipage}{.55\linewidth}
\centering
  \scalebox{.46}{
 \includegraphics[width=2.16in]{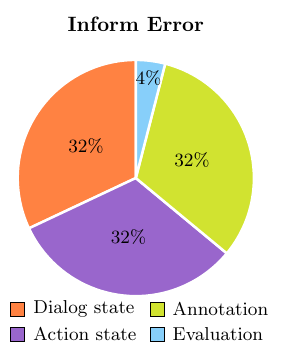}
  }
  \centerline{(b)}\label{fig:inform_b}
\end{minipage}}}
  		\caption{  The domain distribution (a) and primary  reason distribution (b) of inaccurate dialog sessions according to the inform rate metric. The gray bars denote the total number of dialog sessions that contain the corresponding domain; the teal bars denote the number of  dialog sessions with errors in the corresponding domain. }\label{fig:inform}
\end{figure}
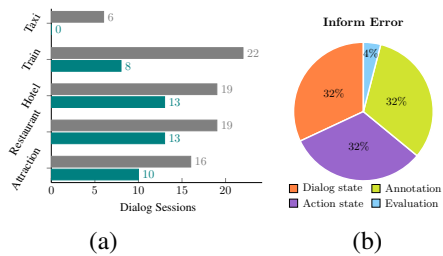 
The detailed domain distribution and primary  reason  distribution of informable slot errors are presented as shown in Figure~\ref{fig:inform}. 
Given that there is no database in the taxi domain, the informable slots are consistently judged to be correct.
The error rate of the dialogs in the hotel and restaurant domains is very high because some informable slots in these two domains are often mispredicted, such as  `\textit{type}' in the hotel domain. As illustrated  in Figure~\ref{fig:inform_b}, 64 percent of dialog informable slot errors are caused by the inaccurate belief states and action states, and the noisy dialog annotations generate 32 percent. 4 percent of that are caused by automatic evaluation scripts and are judged accurately by human evaluation. 
The detailed requestable slot error analysis and 
more examples are provided in  Appendixes~\ref{app:req_error} and~\ref{app:error}, respectively.
In the future, we will focus on solving  errors caused by  the inaccurate dialog/ action states to better apply  Mars-G to real-world scenarios.

\section{Conclusion}
This study explores what dialog context representation is beneficial to improving  task-oriented dialog performance.
Specifically, we propose two contrastive learning strategies to explicitly model the relationship between dialog context and  semantic state representations,  achieving better task completion of a task-oriented dialog system.
Extensive experiments and analysis demonstrate that dialog context representations that are more different from semantic state representations are more beneficial to multi-turn task-oriented dialog.
Moreover, our proposed  Mars achieves state-of-the-art  performance on three datasets.
  
  \section*{Limitations}
The training process of Mars needs to rely on manually annotated belief states and action states as semantic states to explicitly model the relationship between dialog context and semantic state representations through contrastive learning methods. We propose Mars in the research community and hope it can be better applied to real-world scenarios in the industry. However, the annotated data is expensive, which makes our methods have some limitations in the landing process of real scenarios. In the future, to better apply our proposed Mars to real-world scenarios, we will introduce semi-supervised methods to reduce the dependence on annotated dialog corpus. 
\section*{Acknowledgments}
This work was supported by the National Key R\&D Program of China under Grant No. 2020AAA0108600.
\bibliography{custom}

\appendix
\section{Datasets}
\label{sec:dataset}
MultiWOZ 2.0~\cite{budzianowski-etal-2018-multiwoz} is a large-scale English multi-domain task-oriented dialog dataset containing 8438, 1000, and 1000 dialog sessions for training, validation, and testing datasets. It consists of seven domains: attraction, hotel, restaurant, taxi,  train, hospital, and police. 
CamRest676~\cite{wen-etal-2017-network} is a small-scale English restaurant-domain  dataset, which  is split 3/ 1/ 1 for training, validation, and testing datasets. CrossWOZ~\cite{zhu-etal-2020-crosswoz} is a large-scale Chinese multi-domain task-oriented dialog dataset containing  5012, 500, and 500 dialog sessions for training, validation, and testing datasets. It comprises five domains: attraction, restaurant, hotel, taxi, and metro. 
\section{Baselines}
\label{sec:baseline}


Sequicity~\cite{lei-etal-2018-sequicity}, DAMD~\cite{DBLP:conf/aaai/ZhangOY20}, and LABES~\cite{zhang-etal-2020-probabilistic} are copy-augmented GRU-based end-to-end task-oriented dialog systems.
Bidirectional auto-encoding language model BERT~\cite{devlin-etal-2019-bert} is used for the context encoder in DoTS~\cite{DBLP:journals/corr/abs-2103-06648}.
Unidirectional auto-regressive language model GPT-2~\cite{radford2019language} is used in AuGPT~\cite{DBLP:journals/corr/abs-2102-05126},  SOLOIST~\cite{peng2020soloist}, and UBAR~\cite{DBLP:conf/aaai/YangLQ21}. 
Seq2seq language model T5~\cite{2020t5} is used in MinTL~\cite{lin-etal-2020-mintl}, PPTOD~\cite{DBLP:journals/corr/abs-2109-14739},  and MTTOD~\cite{lee-2021-improving-end}. The unified language model UniLM~\cite{DBLP:conf/nips/00040WWLWGZH19} is used in GALAXY~\cite{DBLP:journals/corr/abs-2111-14592}. In addition, auxiliary task-oriented dialog corpora are used to pre-train in AuGPT~\cite{DBLP:journals/corr/abs-2102-05126}, SOLOIST~\cite{peng2020soloist}, PPTOD~\cite{DBLP:journals/corr/abs-2109-14739}, and  GALAXY~\cite{DBLP:journals/corr/abs-2111-14592}.
TRADE~\cite{wu-etal-2019-transferable}, BART-CSP~\cite{DBLP:journals/corr/abs-2111-02574}, and GEEX~\cite{li-etal-2021-generation} are some additional dialog state tracking models. 

\section{States Analysis}
\label{sec:state}
\begin{table}[t]
  \centering
  \scalebox{.78}{
	\begin{tabular}{lccc}
		\toprule
		\bf Model  &\bf Inform & \bf Success & \bf BLEU\\ 
		\midrule
 
End-to-end model &83.2&70.3&19.4\\
\;\;\;\;\;\;\;\;\;\; w/ oracle state&90.8 & 87.4&30.6 \\
      \cdashline{1-4}[1pt/2pt]
  Reference Corpus&     93.7   &  90.9  & 100.0 \\  
		\bottomrule
	\end{tabular}}\caption{Comparison of task-oriented dialog models evaluated on MultiWOZ 2.0. w/ oracle state denotes the system using ground truth belief state and action state for the response generation. Reference results are reported on the official leaderboard of MultiWOZ. \label{tab:pre-experiment}}
\end{table}
To investigate  the impact of belief state and action state on the performance of end-to-end task-oriented dialog, we empirically conduct preliminary experiments on MultiWOZ 2.0~\cite{budzianowski-etal-2018-multiwoz}. 
As shown in Table~\ref{tab:pre-experiment}, the system using ground truth belief state and action state substantially outperforms the traditional end-to-end task-oriented dialog systems, achieving performance  comparable  to reference in terms of  task completion. This demonstrates that the quality of belief state and action state greatly influence on the end-to-end task-oriented dialog performance.
\section{Dialog Context Representation Analysis}
\label{sec:Further}
\begin{figure}[t]
  \centering
  \includegraphics[width=2.8in]{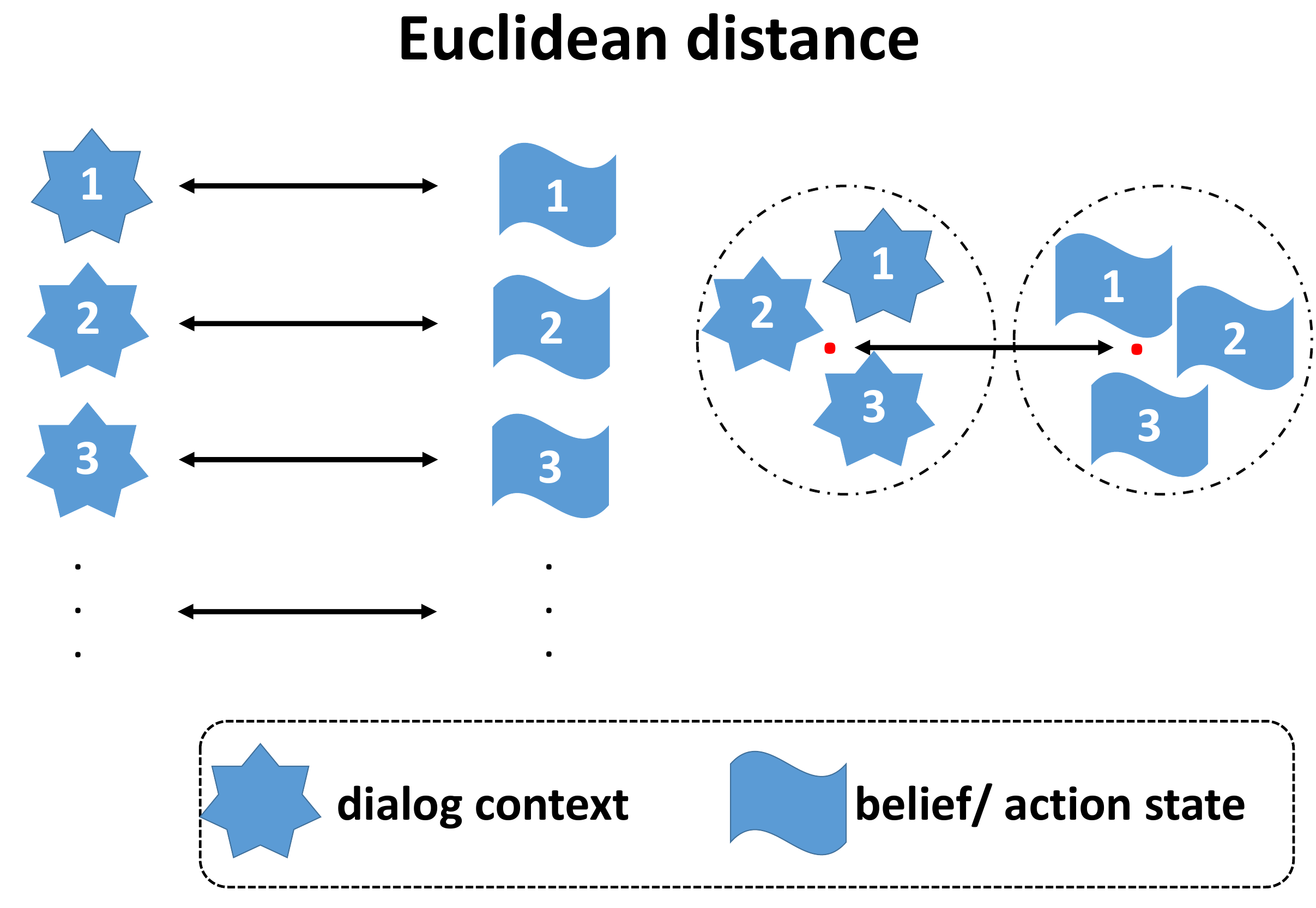}
  \caption{The calculation of Euclidean distance between dialog context and semantic state representations.}
  \label{fig:evaluate}
\end{figure}
To further analyze  dialog context and semantic state representations of Mars-P and Mars-G, we  would measure the similarity of continuous encoder  representation between dialog context and corresponding belief/action state on the MultiWOZ 2.0 test set, as illustrated in Figure~\ref{fig:evaluate}.
Table~\ref{tab:context_1} shows the average L2-normalized  Euclidean distance between dialog context and corresponding  belief/action state representations.
Table~\ref{tab:context_2} shows the  Euclidean distance between the centroids of these two L2-normalized representation spaces. The centroid is the average of all the points in the representation space.
T5 denotes the result before training on the MultiWOZ 2.0.
We find the distance between dialog context and corresponding semantic state representations changes a little before and after training.
Mars-P achieves a smaller distance, thus obtaining a continuous representation of the dialogue context that is semantically more consistent with its semantic state representation.
The distance of Mars-G  is enormous, demonstrating Mars-G  achieves more diverse dialog context representations, different from semantic state representations.

\begin{table}[!t]
  \centering
  \scalebox{.7}{
	\begin{tabular}{lcc}
		\toprule
		\bf Model & \bf Context\&Belief State & \bf Context\&Action State \\ 
		\midrule

T5& 0.797   &   1.018\\   
Baseline& 0.844 &  1.017  \\            
Mars-P&0.340  &  0.542  \\            
Mars-G& 1.996  & 1.993\\              
		\bottomrule
	\end{tabular}}\caption{The distance between dialog context and corresponding semantic state representations on MultiWOZ 2.0.\label{tab:context_1}}
\end{table}

\begin{table}[!t]
  \centering
  \scalebox{.7}{
	\begin{tabular}{lcc}
		\toprule
		\bf Model & \bf Context\&Belief State & \bf Context\&Action State \\ 
		\midrule

T5& 0.555  &  0.807  \\ 
Baseline& 0.598  &   0.699  \\  
Mars-P&  0.042  &  0.046  \\
Mars-G& 1.993 & 1.987\\ 
		\bottomrule
	\end{tabular}}\caption{The distance between the centroids of these two  representation spaces on MultiWOZ 2.0.\label{tab:context_2}}
\end{table}

\section{Hyper-parameter Analysis}
\label{sec:parameter}
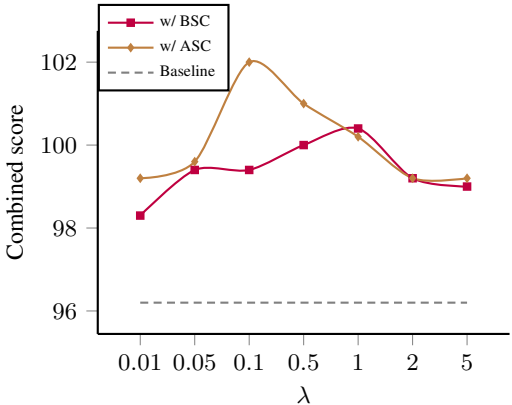
\begin{figure}[!t]
\setlength{\abovecaptionskip}{0pt}
\begin{center}
\pgfplotsset{height=5.6cm,width=8.5cm,compat=1.15,every axis/.append style={thick}}
\begin{tikzpicture}
\tikzset{every node}=[font=\small]
\begin{axis}
[width=7cm,enlargelimits=0.13, tick align=outside, legend style={cells={anchor=west},legend pos=south west, legend columns=1,every axis legend/.append style={at={(0,0.8)}}}, xticklabels={ $0.01$, $0.05$, $0.1$, $0.5$,$1$,$2$,$5$},
xtick={0,1,2,3,4,5,6},
		axis y line*=left,
		axis x line*=left,
ylabel={Combined score},xlabel={$\lambda$},font=\small]

\addplot+ [sharp plot, solid,mark=square*,mark size=1.2pt,mark options={mark color=purple}, color=purple][smooth] coordinates
{  (0,98.3) (1,99.4) (2,99.4) (3,100) (4,100.4)(5,99.2)(6,99)};
\addlegendentry{\tiny w/ BSC}

\addplot+ [sharp plot,solid, mark=diamond*,mark size=1.2pt,mark options={mark color=brown}, color=brown][smooth] coordinates
{  (0,99.2) (1,99.6) (2,102) (3,101) (4,100.2)(5,99.2)(6,99.2)};
\addlegendentry{\tiny {w/ ASC}}

\addplot+ [sharp plot,densely dashed,no markers, color=gray] coordinates
{ (0,96.2) (1,96.2) (2,96.2) (3,96.2) (4,96.2) (5,96.2)(6,96.2)};
\addlegendentry{\tiny Baseline}
\end{axis}
\end{tikzpicture}
\caption{\label{fig:lambda}The  Mars-G performance with different levels of hyper-parameter $\lambda$ on the MultiWOZ 2.0. w/ BSC denotes belief state  module, w/ ASC denotes action state  module. $T$ is set to 0.5.}

\end{center}
\end{figure}

\begin{figure}[!t]
\setlength{\abovecaptionskip}{0pt}
\begin{center}
\pgfplotsset{height=5.6cm,width=8.5cm,compat=1.15,every axis/.append style={thick}}
\begin{tikzpicture}
\tikzset{every node}=[font=\small]
\begin{axis}
[width=7cm,enlargelimits=0.13, tick align=outside, legend style={cells={anchor=west},legend pos=south west, legend columns=1,every axis legend/.append style={at={(0,0.75)}}}, xticklabels={ $0.01$, $0.1$, $0.5$,$1$,$5$,$10$},
xtick={0,1,2,3,4,5,6},
		axis y line*=left,
		axis x line*=left,
ylabel={Combined score},xlabel={$T$},font=\small]

\addplot+ [sharp plot, solid,mark=square*,mark size=1.2pt,mark options={mark color=purple}, color=purple][smooth] coordinates
{  (0,97.6) (1,99.4) (2,100.4) (3,99.71) (4,99.1)(5,99)};
\addlegendentry{\tiny w/ BSC}

\addplot+ [sharp plot,solid, mark=diamond*,mark size=1.2pt,mark options={mark color=brown}, color=brown][smooth] coordinates
{  (0,98.8) (1,99.8) (2,102) (3,99.3) (4,98.7)(5,98.1)};
\addlegendentry{\tiny {w/ ASC}}

\addplot+ [sharp plot,densely dashed,no markers, color=gray] coordinates
{ (0,96.2) (1,96.2) (2,96.2) (3,96.2) (4,96.2) (5,96.2)};
\addlegendentry{\tiny Baseline}
\end{axis}
\end{tikzpicture}
\caption{\label{fig:T}The  Mars-G performance with different levels of hyper-parameter $T$ on the MultiWOZ 2.0. $\lambda_1$ is set to 1 for  w/ BSC, and $\lambda_2$ is set to 0.1 for w/ ASC.}

\end{center}
\end{figure}
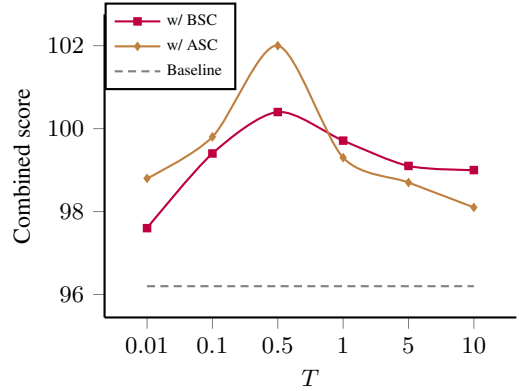

We empirically investigate how the hyper-parameters $\lambda$ and $T$ for both modules of Mars-G affect the  performance of task-oriented dialog on the MultiWOZ 2.0, respectively. The selection of $\lambda$ influences the role of the contrastive learning objective function across the entire task-oriented dialog training process.  As   Figure \ref{fig:lambda} shows, $\lambda$ ranging from 0.01 to 5 nearly all improve task-oriented dialog performance. This indicates our proposed Mars-G is robust and effective.
When  $\lambda =0.1$, w/ ASC achieves the best performance.
When  $\lambda =1$, w/ BSC achieves the best performance.
The selection of $T$ affects   the differentiation of  hard negative samples. The smaller the value of $T$ is, the more attention is paid to distinguishing complex  negative samples.
As shown in Figure \ref{fig:T}, combined scores increase for almost all $T$ values ranging from 0.01 to 10, and the best performance is achieved when $T =0.5$ for both modules of Mars-G.

 

\section{Visualization}
\label{app:vis}
\begin{figure*}[ht]
  \centering
  \includegraphics[width=6.2in]{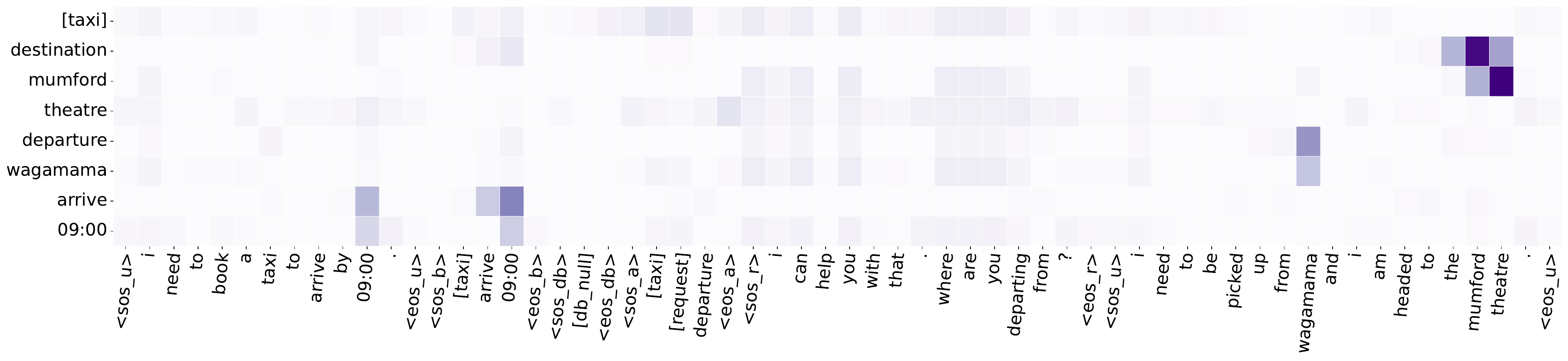}
  \caption{Visualization of the cross-attention weights between dialog context and generated belief states for our proposed Mars-G. The horizontal axis is the dialog context, and the vertical axis is the generated belief state.}
  \label{bs_vis}
\end{figure*}
\begin{figure*}[ht]
  \centering
  \includegraphics[width=6.2in]{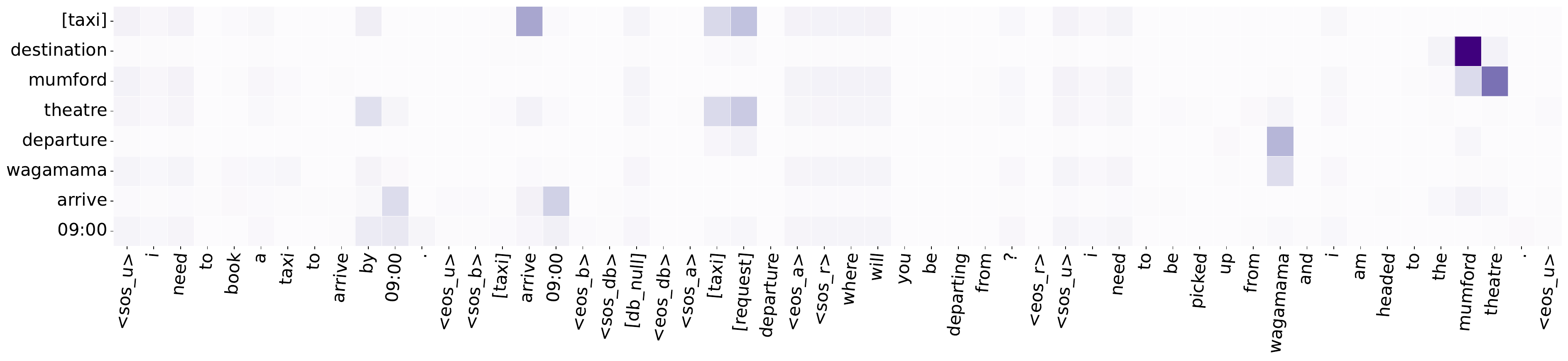}
  \caption{Visualization of the cross-attention weights between dialog context and generated belief states for the baseline system.}
  \label{base_bs_vis}
\end{figure*}
We provide an example to visualize the   dialog state tracking process of our proposed Mars-G and baseline system. The cross-attention weights between dialog context and generated belief states from the last layer of the transformer decoder stack  are shown in Figures~\ref{bs_vis} and ~\ref{base_bs_vis}.  Compared with the baseline system,  Mars-G could achieve more accurate attention weights.
The slot `\textit{arrive 09:00}' assigns high attention weights for the user utterance `\textit{09:00}' and previous belief state `\textit{arrive 09:00}'. Similarly, the slots `\textit{destination mumford theatre}' and  `\textit{departure wagamama}' accurately give  high attention weights for the corresponding user utterance.   The visualization further demonstrates that Mars-G could achieve  more reasonable dialog context representation to generate accurate belief states.

\section{Further Ablation Analysis}
\label{app:sim}

\begin{table}[!t]
  \centering
  \scalebox{.75}{
	\begin{tabular}{lcccc}
		\toprule
		\bf Model & \bf Inform & \bf Success & \bf BLEU & \bf Combined\\ 
		\midrule

Baseline& 83.2&70.3&19.4&96.2\\
Mars-variant& 85.7 &74.8 &19.6  &99.9\\
Mars-P&86.6&75.5&19.6&100.7\\
		\bottomrule
	\end{tabular}}\caption{The performance of the different methods  on MultiWOZ 2.0. Mars-variant denotes similarity strategy.\label{tab:ablation2}}
\end{table}
To  get a more complete picture of the effectiveness of Mars-P, we introduce a  similarity strategy (Mars-variant). We use the cosine similarity function to narrow the distance between the continuous representation of dialog contexts and semantic states for the same dialog session to model the relationship between dialog context and corresponding semantic state representations. We don't distinguish the continuous representation of  dialog context and states for different dialog sessions.
As shown in Table~\ref{tab:ablation2}, Mars-variant outperforms the  baseline system  by  3.7 combined scores, indicating the effectiveness of the relationship modeling between dialog context and corresponding semantic representations. In addition, Mars-variant underperforms
Mars-P by  0.8 combined scores. This demonstrates that  distinguishing the continuous representation of  dialog context and states for different dialog sessions  is beneficial for dialog modeling.
\section{Low Resource Scenario Results}
\label{app:low}
We train all dialog systems five times with different random seeds in the low resource scenario. The detailed results of 5 runs are provided in Table~\ref{tab:low_resource_detail}.

\begin{table*}[t]
   \small
   \centering
   \scalebox{.62}{
	\begin{tabular}{lcccccccccccccccc}
		\toprule
		\multirow{2}*{\bf Model} &\multicolumn{4}{c}{\bf 5\% } &\multicolumn{4}{c}{\bf 10\% } &   \multicolumn{4}{c}{\bf 20\%}&  \multicolumn{4}{c}{\bf 50\% }\\ 
\cmidrule(lr){2-5} \cmidrule(lr){6-9} \cmidrule(lr){10-13} \cmidrule(lr){14-17} 
		& \bf Inform  & \bf Success & \bf BLEU & \bf Combined& \bf Inform  & \bf Success & \bf BLEU& \bf Combined& \bf Inform  & \bf Success & \bf BLEU& \bf Combined& \bf Inform  & \bf Success & \bf BLEU& \bf Combined\\ 
		\midrule

		\multicolumn{17}{c}{\bf DAMD }\\
			\midrule
			
run 1& 35.4 &  17.2& 10.9 &  37.2  &  41.3 & 23.7 & 12.4  & 44.9   &  51.8 &  31.9 &  14.1 &  56.0 &  60.1  &  44.2 & 15.6  &  67.8\\
run 2& 40.8 &  20.9 &  12.0 & 42.9  & 41.5 & 25.3& 11.2& 44.6 &50.4&  32.4 & 13.8  & 55.2&54.7  &  39.5 &  14.8& 61.9 \\
run 3& 38.5 &  14.5&  10.6  &  37.1&  40.0& 23.9&12.3&  44.3  & 42.5& 26.8 & 15.4& 50.1 &59.1 &   45.7 &   15.1&67.5\\
run 4&35.4 &  16.5&10.2 & 36.2  &  42.3&  20.2& 12.0 &43.3  &46.1& 29.2&  14.0 & 51.7 &57.2& 43.0&16.1&  66.2  \\
run 5& 34.0& 17.6& 12.4&  38.2  & 39.5  & 21.9  & 13.0  & 43.7  &  50.8 &31.3  & 13.6  & 54.7& 63.1  &   49.3& 17.1  &   73.3                   \\

\cdashline{1-17}[1pt/2pt]
Average&   36.8&   17.3&  11.2&   38.3& 40.9  &23.0   & 12.2  &    44.2&  48.3 & 30.3 &   14.2&  53.5&  58.8    & 44.3  &15.7    &67.3     \\

		\midrule
		\multicolumn{17}{c}{\bf MinTL }\\
			\midrule  
run 1& 54.4  & 41.1  &  14.2 & 62.0  & 55.8 & 44.0 &15.3& 65.2& 62.5& 54.2&17.3&75.7& 71.7  & 62.7  & 16.9  &84.1\\
run 2&  54.8  & 36.8 & 13.6 & 59.4  & 51.6 & 42.0 & 15.7&62.5 &      65.8  &  56.3   &   15.5  &  76.6  & 67.4  &   59.7& 18.7  & 82.3  \\
run 3& 53.3&39.3&14.2&60.5       &   55.1       &  44.7  & 16.1& 66.0 &  68.0 &  59.0  & 16.6  &  80.1 & 70.6  &  62.6 &  17.5  &84.1\\
run 4& 52.4&37.1&13.8&58.6 & 58.4 & 47.3& 15.2 & 68.1    &   58.3 & 48.4  & 14.4  &67.8& 68.9&  61.3 & 18.2 &  83.3\\
run 5& 47.5  &  36.3 & 13.8& 55.7  & 56.8 & 46.4 & 15.9 & 67.5 & 66.9 & 56.8  & 17.0  & 78.9  &73.1&  64.5 &  18.6 & 87.4\\
\cdashline{1-17}[1pt/2pt]
Average&  52.5 & 38.1 & 13.9 &59.2  &55.5  & 44.9  & 15.6 &65.8  &  64.3 & 54.9  &  16.2& 75.8 & 70.3&  62.2 & 18.0  &  84.3 \\
	\midrule
		\multicolumn{17}{c}{\bf UBAR }\\
			\midrule
run 1& 37.4  &  23.0 &  11.6 & 41.8  &52.3&34.8  & 13.0  & 56.6& 61.7   &   45.7 & 15.9   &  69.6  &  77.2 &   61.5 & 15.5 &  84.9\\
run 2&  33.3  & 20.6  & 11.2  & 38.2  &  48.5 & 35.9& 14.5 & 56.7 &63.4& 47.8& 15.5 &71.1 & 78.0  &  63.8 &16.9  & 87.8  \\

run 3& 40.0   &  23.1 & 11.7  & 43.3 &50.3&33.2&13.6&  55.4&67.8   &   50.0&  13.1 & 72.0  & 77.4  & 64.6 &  16.2 &  87.2\\

run 4& 38.2  & 22.4  &   10.7 &  41.0  & 52.5& 34.6 & 12.5&  56.1 &  68.3 & 51.7  & 14.4  & 74.4&  78.5&  64.1 & 16.8  &  88.1 \\
run 5&38.0 &  21.3 & 11.3  &  41.0 &  47.8 &  32.3&13.7   & 53.8  &  66.2  &  48.3   &   13.8 &71.1&76.8&  62.4 & 16.2   &   85.8\\
\cdashline{1-17}[1pt/2pt]
Average&  37.4 & 22.1 & 11.3  & 41.1  &  50.3& 34.2&  13.5 &55.8 & 65.5  & 48.7  &14.5   &71.6   &77.6   & 63.3  & 16.3  & 86.8 \\
	\midrule
	
				\multicolumn{17}{c}{\bf MTTOD }\\
			\midrule
			
run 1&51.4& 37.5&12.0& 56.5& 70.9& 58.0& 13.8&  78.3&  71.1 &  59.0 & 14.2  &  79.3&   74.7&  64.4& 15.2 & 84.8\\
run 2&  53.8  & 41.7  & 11.3 & 59.1   &64.1    &   53.7 & 13.8  & 72.7  & 69.5   &  60.7 &  14.0 &  79.1 & 79.3  & 67.7  & 15.0  &  88.5  \\
run 3& 55.7  & 31.1  & 11.5  & 54.9  & 61.0&  50.8&13.7&69.6 & 78.4 & 65.1& 14.7 & 86.5&  82.3 &  71.1 &   15.5& 92.2  \\
run 4& 52.4  & 33.3  & 10.6  &  53.5  &  73.0 &   59.3& 14.0 & 80.2& 80.2   & 67.4  &  14.5  &  88.3  &            76.6   &  65.6 &  15.3  &   86.4 \\
run 5&  58.0 &  43.2 &  11.3 & 61.9  &  65.4 &  54.2 & 13.7  & 73.5  &  75.9 & 64.3  &  14.1&84.2&  79.8 &  68.7 &  15.1  &  89.4  \\
\cdashline{1-17}[1pt/2pt]
Average& 54.3&37.4& 11.3  & 57.2&   66.9&  55.2 & 13.8  &  74.9  & 75.0& 63.3& 14.3    &  83.5 & 78.5 &  67.5 &  15.2  &  88.2  \\
	\midrule
			\multicolumn{17}{c}{\bf PPTOD }\\
			\midrule
			
run 1& 70.7& 46.8& 13.7&72.5   & 65.2& 50.6&14.2&72.1&    72.3     &  55.0  &   14.9&78.6&    74.8&  60.4 & 15.8  &83.4\\
run 2&  64.6& 45.8 &  13.8 &  69.0 & 69.3  &  52.9 & 15.3  & 76.4&70.5   &  57.7 &  17.7  & 81.8  &74.1 & 64.2 & 16.4 & 85.6\\
run 3& 64.4 &  51.1 &  15.1 & 72.9 &  65.7 &  53.6 & 15.8  & 75.5   & 74.8  &  64.6 & 16.9  & 86.6   &  74.3 &  61.8 &  17.2 &  85.3 \\
run 4& 63.9  & 47.0  &  14.7 & 70.2  &  70.1& 55.4  &  17.8 & 80.6  & 71.8  & 57.3  &  16.0  &  80.6  &   76.4 & 63.7  &  18.0&   88.1\\
run 5& 63.7  & 50.7 &  14.4 &  71.6 & 71.2  & 55.8  &  15.6 &   79.1&  74.1 &  61.6 & 15.8  &  83.7 &  74.4 &   61.9&  17.5  & 85.7 \\
\cdashline{1-17}[1pt/2pt]
Average&      65.5&     48.3    &14.3&  71.2       &   68.3 &  53.7& 15.7  &76.7& 72.7& 59.2&16.3&82.3&  74.8 &  62.4& 17.0&85.6 \\

	\midrule

		\multicolumn{17}{c}{\bf Mars-G }\\
			\midrule

run 1& 55.8&41.1&14.0&   62.5  &  68.7 &   55.0& 16.7     &    78.6&   72.4&60.2    &   18.1&  84.4&  82.6 & 70.2  &  18.8 &  95.2\\
run 2&  57.0     &  43.2 &  12.9 &  63.0  &    68.4      & 55.9 & 15.2  &  77.4 &       76.0     &  61.4 &17.1  & 85.8  & 78.4  & 66.9 & 18.7  &91.4    \\
run 3& 61.4  & 46.7  &  14.5 & 68.6  &  68.9 & 53.8  & 14.0 &  75.4 & 76.6  & 63.8  &17.0   &     87.2      & 82.8  & 73.6  &  17.9   &      96.1   \\
run 4& 56.1  & 42.4  & 14.1   & 63.4  & 73.1 &  60.3 &  16.6 & 83.3  &  80.6 & 63.9 & 17.1  &      89.4    &   82.5&  71.3 & 19.0      &95.9   \\
run 5& 57.8  & 43.5  & 13.8  & 64.5   & 67.7  &  51.5&  15.7 &  75.3 &     77.7       &  65.0 & 16.8 & 88.2  &  84.6 &   74.2 & 18.7 &  98.1 \\

\cdashline{1-17}[1pt/2pt]
Average& 57.6  & 43.4 & 13.9 & 64.4& 69.4 & 55.3  & 15.6 & 78.0 & 76.7  &  62.9 & 17.2 &  87.0&  82.2& 71.2 & 18.6 &95.3  \\
	
		\bottomrule
	\end{tabular}}\caption{Comparison of task-oriented dialog systems on the MultiWOZ 2.0 in the low resource scenarios.\label{tab:low_resource_detail}}

\end{table*}
\begin{figure}[!t]
  \centering\scalebox{.88}{
    \subfigure{
  \begin{minipage}{.4\linewidth}
  \centering\scalebox{.46}{
\begin{tikzpicture}
\begin{axis}
[
    xbar, enlarge y limits=0.1,
enlarge x limits={0.1,upper},
    xlabel={\ Dialog Sessions}, symbolic y coords={  Attraction,Restaurant,Hotel,Train,Taxi}, 
    ytick=data, nodes near coords, 
    nodes near coords align={horizontal},axis y line*=left,
		axis x line*=left, yticklabel style={rotate=60},]
\addplot  +[color=brown] coordinates { (7,Train) (8,Restaurant) (11,Hotel) (13,Attraction) (1,Taxi)};
\addplot +[color=gray] coordinates {(18,Train) (17,Restaurant) (22,Hotel) (15,Attraction) (9,Taxi)};
\end{axis}
\end{tikzpicture}}  \centerline{(a)}

  \end{minipage}}
    \subfigure{
\begin{minipage}{.55\linewidth}
\centering
  \scalebox{.46}{

 \includegraphics[width=2.16in]{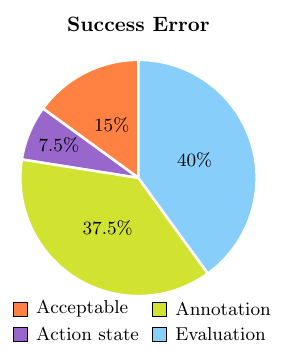}

}  \centerline{(b)}\label{fig:success_b}
\end{minipage}}}
	\caption{\label{fig:success} The domain distribution (a) and primary  reason distribution (b) of inaccurate dialog sessions according to  success rate metric. The gray bars denote the total number of dialog sessions that contain the corresponding domain; the brown bars denote the number of  dialog sessions with errors in the corresponding domain.}
\end{figure}
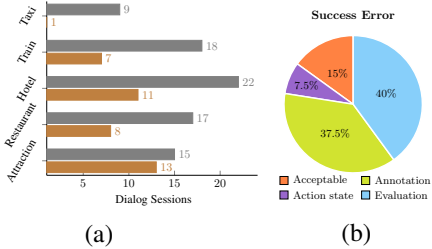 
\section{Requestable Slot Error Analysis}
\label{app:req_error}
Considering the inclusion relationship of the two metrics described in Section~\ref{sec:evaluation}, we  select dialog sessions with the wrong success rate and accurate inform rate for success rate error analysis.
The detailed domain distribution and primary  reason  distribution of requestable slot errors are presented as shown in Figure~\ref{fig:success}.  The error rate of the dialogs in the taxi and train domains is very low because requestable slots in these two domains are few and simple.  For example, the requestable slot in the taxi domain only has `\textit{phone}'. The error rate of the dialogs in the attraction domain is very high. 
As illustrated  in Figure~\ref{fig:success_b}, 77.5 percent of dialog requestable slot errors are caused by  the noisy dialog annotations and automatic evaluation scripts. 15 percent of generated system responses are acceptable.
When users request some information about something and do not ask for a specific requestable slot, Mars-G generates system responses that lack some requestable slots such as `\textit{postcode}' and `\textit{address}'. In addition, Mars-G requests users some other useful information instead of providing booked reference directly. We think system responses generated by Mars-G in both cases are reasonable.
Inaccurate action states cause 7.5 percent of dialog requestable slot errors.

\section{Examples for Error Analysis}
\label{app:error}

Tables \ref{tab:case1} - \ref{tab:case9} show several examples generated by Mars-G for detailed error analysis. 
As shown in Table \ref{tab:case1},  Mars-G generates the inaccurate belief state `\textit{food jamaican}' rather than `\textit{food italian}', leading to the informable slot error. 
 Table \ref{tab:case2} shows that  Mars-G generates the inadequate  action state,  not including the slot name `\textit{name}', leading to the informable slot error. 
Table \ref{tab:case3} shows that the informable slot error is caused by automatic evaluation. Mars-G provides the accurate response in turn 7. However, the automatic evaluation script estimates the wrong active domain `\textit{[taxi]}' rather than `\textit{[attraction]}' from the belief state.
The informable slot error in Table~\ref{tab:case4}  is caused by noisy dialog annotations. The informable slot `pricerange moderate' does not appear in the conversation.

As shown in Table \ref{tab:case5},  Mars-G generates the inaccurate action state `\textit{[request] people}' provided in the belief state `\textit{people 1}', leading to the requestable slot error. 
Table \ref{tab:case6} shows that the requestable slot error is caused by automatic evaluation.
Mars-G provides the accurate response in turn 2, while the automatic evaluation script only determines if it offers a reference in turn 3.
We think system responses generated by Mars-G in Tables~\ref{tab:case7} and ~\ref{tab:case8} are acceptable.
When users request some information about something and do not ask for a specific requestable slot, Mars-G generates system responses, lack of the  requestable slot `\textit{postcode}', as shown in Table~\ref{tab:case7}. In addition, Mars-G requests users whether to make a reservation instead of providing booked references directly, as shown in Table~\ref{tab:case8}.
Noisy dialog annotations cause the requestable slot error in Table~\ref{tab:case9}. The requestable slot `postcode' does not appear in the conversation.

\begin{table*}[ht]
  \centering
  \scalebox{.8}{
	\begin{tabular}{ll}
		\toprule
Goal:& [restaurant]: \{informable: {\color{red}food italian} area east\} \{requestable: reference\}\\

		\midrule
 User (Turn 0):& i am looking for a {\color{red}jamaican} restaurant in the east.\\
Golden R: &there are no [value\_food] restaurants in the [value\_area].\\ &would you like to try another food type or area?\\
Golden BS: & [restaurant] {\color{red}food jamaican} area east\\
Golden AS:&   [restaurant] [nooffer] food area [request] area food      \\
 Mars-G R:& i am sorry, there are no [value\_food] restaurants in the [value\_area]. \\&would you like to try another area or another type of food?\\
Mars-G BS: &[restaurant] {\color{red}food jamaican} area east\\
  Mars-G AS:&     [restaurant] [nooffer] food area [request] area food   \\
 
 	\midrule
User (Turn 1): & i am looking for a place that serves {\color{red}jamaican food} in the east. if not, {\color{red}italian} will do.	\\
   Golden R: & there is [value\_choice] [value\_food] place in the [value\_area], [value\_name]. 	\\
 Golden BS: &  [restaurant] {\color{red}food italian} area east	\\
 Golden AS:&     [restaurant] [inform] choice name food area   \\
  Mars-G R:  &i am sorry, there are no [value\_food] restaurants in the [value\_area]. \\&would you like to try another type of food?\\
  Mars-G BS: & [restaurant] {\color{red}food jamaican} area east\\
  Mars-G AS:&    [restaurant] [nooffer] food area [request] food    \\
 
  	\midrule
 User (Turn 2):  & ok, i would like to make a reservation there.	\\
  	Golden R: & i can do that for you, i just need details such as day, time, and the size of your party. \\
  	Golden BS: &  [restaurant] {\color{red}food italian} area east\\
Golden AS:&    [restaurant] [offerbook] [request] time day people    \\
  	  Mars-G R: &  what day and time would you like to dine? and how many people will be in your party?\\
  	 Mars-G BS: & [restaurant] {\color{red}food jamaican} area east\\
  Mars-G AS:&    [restaurant] [request] day time people    \\

  	 	\midrule
  	 	User (Turn 3):   &i need a table for 7 at 14:45 on friday.\\
  	 	  Golden R: & i booked you a table for [value\_people] on [value\_day] at [value\_time] at \\&[value\_name], reference: [value\_reference]. can i assist you further today?\\
  	 	 	Golden BS:  & [restaurant] {\color{red}food italian} area east time 14:45 day friday people 7 \\
  	 	 	Golden AS:&    [restaurant] [offerbooked] time reference name day people [general] [reqmore]    \\
  	 	     Mars-G R: & booking was successful. the table will be reserved for 15 minutes. \\&reference number is: [value\_reference].\\
  	 	 Mars-G BS:  & [restaurant] {\color{red}food jamaican} area east time 14:45 day friday people 7 \\
  Mars-G AS:&    [restaurant] [offerbooked] reference    \\
    
		\bottomrule
	\end{tabular}}\caption{An example in dialog session pmul0367. The informable slot error is caused by the inaccurate belief state. R denotes response, BS denotes belief state, and AS denotes action state. \label{tab:case1}}
\end{table*}

\begin{table*}[ht]
  \centering
  \scalebox{.8}{
	\begin{tabular}{ll}
		\toprule
Goal:& [attraction]: \{informable: area centre\} \\

		\midrule
 User (Turn 3):& i am also looking for places to go in town. perhaps an attraction in the city centre.\\
Golden R: &i have the {\color{red}[value\_name]} located at [value\_address] and it's [value\_price] entrance.\\
Golden BS: &[attraction] area centre\\
Golden AS:&   [attraction] [recommend] address price {\color{red}name}     \\
 Mars-G R:& there are [value\_choice] attractions in the city centre. do you have a preference for type?\\
Mars-G BS: &[attraction] area centre  \\
  Mars-G AS:&    [attraction] {\color{red}[inform] choice} [request] type  \\
 
 	\midrule
User (Turn 4): &	that sounds perfect. thanks!\\
   Golden R: & can i help you with anything else?	\\
 Golden BS: &  [attraction] area centre	\\
 Golden AS:&  [general] [reqmore]    \\
  Mars-G R:  &is there anything else i can help you with?\\
  Mars-G BS: & [attraction] area centre \\
  Mars-G AS:&    [general] [reqmore] \\
		\bottomrule
	\end{tabular}}\caption{An example in dialog session pmul2563. The informable slot error is caused by the inaccurate action state.
\label{tab:case2}}
\end{table*}

\begin{table*}[ht]
  \centering
  \scalebox{.8}{
	\begin{tabular}{ll}
		\toprule
Goal:& [attraction]: \{informable: area south\} \{requestable: postcode\}\\

		\midrule
 User (Turn 4):& i would also like to go to some attractions in the same area as the restaurant.\\
Golden R: & there are [value\_choice] attractions in the area. there's a [value\_type], [value\_type],\\& [value\_type], [value\_type], [value\_type]... do any of these appeal to you in particular?\\
Golden BS: &[attraction] area south [restaurant] food indian pricerange expensive area south\\
Golden AS:&   [attraction] [select] type [inform] choice     \\
 Mars-G R:&  there are [value\_choice] attractions in the area. do you have a preference for type?\\
Mars-G BS: &[attraction] area south [restaurant] food indian pricerange expensive area south\\
  Mars-G AS:& [attraction] [inform] choice [request] type       \\
 
 	\midrule
User (Turn 5): &	restaurant is what i am looking for, entrance fee and postcode please. {\color{red}get me a taxi }\\&{\color{red}from the attraction by 10.15, get me its contact number and car type.}\\
   Golden R: & just to clarify you are looking for another restaurant and not an attraction?	\\
 Golden BS: &  [attraction] area south [restaurant] food indian pricerange expensive area south	\\
 Golden AS:&    [attraction] [select] [restaurant] [select]  \\
  Mars-G R:  &[value\_name] is located in postcode [value\_postcode]. the phone number is \\&[value\_phone]. the entrance fee is [value\_price]. can i help you with anything else?\\
  Mars-G BS: & {\color{red}[taxi]} leave 10.15 destination tandoori palace departure the nirala [attraction] \\&area south [restaurant] food indian pricerange expensive area south \\
  Mars-G AS:&    [attraction] [inform] price phone name postcode [general] [reqmore] \\
 
  	\midrule
 User (Turn 6):  & oh, yeah, i do want information on attractions! in the same area. after that, i will need a taxi.	\\
  	Golden R: & if you like parks the [value\_name] has [value\_price] admission and is in the \\&[value\_postcode] postcode. or i can tell you about the other attractions? \\
  	Golden BS: &  [attraction] area south [restaurant] food indian pricerange expensive area south\\
Golden AS:&     [attraction] [recommend] name postcode price [general] [reqmore]  \\
  	  Mars-G R: & there are [value\_choice] attractions in the [value\_area]. [value\_choice] are \\&[value\_type], [value\_choice] are [value\_type], and [value\_choice] is [value\_type].\\& do you have a preference?\\
  	 Mars-G BS: & [taxi] leave 10.15 destination tandoori palace departure the nirala [attraction] \\&area south [restaurant] food indian pricerange expensive area south \\
  Mars-G AS:&  [attraction] [inform] area type choice [request] type    \\

  	 	\midrule
  	 	User (Turn 7):   &that sounds wonderful, is it also in the south part of town? {\color{red}if so, i would like a taxi}\\& {\color{red}from there to my hotel.}\\
  	 	  Golden R: & yes, [value\_name] is in the [value\_area]. in order to help you book a taxi between \\&the park and your hotel, i need to know what hotel you're at.\\
  	 	 	Golden BS:  & [taxi] departure wandlebury country park [attraction] area south \\&[restaurant] food indian pricerange expensive area south \\
  	 	 	Golden AS:&    [attraction] [inform] area name [taxi] [request] destination  \\
  	 	     Mars-G R: &{\color{red}yes, [value\_name] is in the [value\_area].}\\
  	 	 Mars-G BS:  &{\color{red}[taxi]} leave 10.15 destination tandoori palace departure the nirala [attraction] \\&area south [restaurant] food indian pricerange expensive area south \\
  Mars-G AS:&  [attraction] [inform] area name   \\
    
		\bottomrule
	\end{tabular}}\caption{An example in dialog session mul2491. The informable slot error is caused by automatic evaluation.\label{tab:case3}}
\end{table*}

\begin{table*}[ht]
  \centering
  \scalebox{.8}{
	\begin{tabular}{ll}
		\toprule
Goal:& [restaurant]: \{informable: {\color{red}pricerange moderate} area west \}\{requestable: reference\}\\

		\midrule
 User (Turn 0):& hi! i am planning a trip in cambridge and i am looking for a restaurant in the west part of town.\\
Golden R: &there are [value\_choice] restaurants in the [value\_area]. is there any kind of food you're looking for?\\
Golden BS: &{\color{red}[restaurant] area west}\\
Golden AS:&    [restaurant] [inform] choice area [request] food    \\
 Mars-G R:& there are [value\_choice] restaurants in the [value\_area] part of town.\\& do you have a price range or type of food in mind?\\
Mars-G BS: &{\color{red}[restaurant] area west}\\
  Mars-G AS:&       [restaurant] [inform] choice area [request] food price \\
 
 	\midrule
User (Turn 1): &	how about you suggest one a book a table for 4 on tuesday at 15:45 and \\&i need the reference number too please\\
   Golden R: & i was able to book [value\_name], reference number is [value\_reference]	\\
 Golden BS: &  {\color{red} [restaurant] area west time 15:45 day tuesday people 4}	\\
 Golden AS:&  [restaurant] [offerbooked] name reference    \\
  Mars-G R:  &i have booked you at [value\_name]. your reference number is [value\_reference]. \\&is there anything else i can help you with?\\
  Mars-G BS: & {\color{red}[restaurant] area west time 15:45 day tuesday people 4}\\
  Mars-G AS:&   [restaurant] [offerbooked] reference name [general] [reqmore]  \\

		\bottomrule
	\end{tabular}}\caption{An example in dialog session pmul3992. The informable slot error is caused by  noisy dialog annotations.\label{tab:case4}}
\end{table*}

\begin{table*}[ht]
  \centering
  \scalebox{.8}{
	\begin{tabular}{ll}
		\toprule
Goal:& [train]: \{informable: leave 17:00 destination birmingham new street  day monday departure cambridge\}\\
&    \;\;\;\;\;\;\;\;\;\; \{requestable:  {\color{red}reference}\}
\\
		\midrule
 User (Turn 0):&what train do i take to get to birmingham new street? i need to leave after 17:00. \\
Golden R: &there are trains departing hourly at 1 minute past the hour [value\_leave]. all are [value\_price], \\&and take [value\_time].\\
Golden BS: &[train] destination birmingham new street leave 17:00\\
Golden AS:&    [train] [inform] price leave time   \\
 Mars-G R:& where will you be departing from and what day would you like to travel ?\\
Mars-G BS: &[train] destination birmingham new street leave 17:00\\
  Mars-G AS:&    [train] [request] departure day  \\

  	 	\midrule
  	 	User (Turn 1):   &i am looking to travel on monday. the first train at 17:01 would be fine.  can i get that train info?\\
  	 	  Golden R: & sure the trainid is [value\_id]. it leaves at [value\_leave] and arrives at [value\_arrive].\\& would you like to make a reservation?\\
  	 	 	Golden BS:  & [train] destination birmingham new street day monday departure cambridge leave 17:00 \\
  	 	 	Golden AS:&   [train] [inform] arrive leave id [offerbook]  \\
  	 	     Mars-G R: &[value\_id] leaves at [value\_leave] and arrives at [value\_arrive]. would you like me to book that for you?\\
  	 	 Mars-G BS:  & [train] destination birmingham new street day monday departure cambridge leave 17:00\\
  Mars-G AS:&   [train] [inform] arrive leave id [offerbook]   \\
   	 	\midrule
  	 	User (Turn 2):   &yes please make a reservation and give me the reference number.\\
  	 	  Golden R: & your reservation is book. the reference number is [value\_reference].\\& can i assist you with anything else today?\\
  	 	 	Golden BS:  &  [train] destination birmingham new street day monday departure cambridge people 1 leave 17:00 \\
  	 	 	Golden AS:&   [train]  {\color{red}[offerbooked] reference} [general] [reqmore]  \\
  	 	     Mars-G R: &how many tickets do you need?\\
  	 	 Mars-G BS:  &[train] destination birmingham new street day monday departure cambridge  {\color{red}people 1} leave 17:00 \\
  Mars-G AS:&    [train]  {\color{red}[request] people}  \\   
		\bottomrule
	\end{tabular}}\caption{An example in dialog session mul1624. The inaccurate action state causes the requestable slot error. \label{tab:case5}}
\end{table*}

\begin{table*}[ht]
  \centering
  \scalebox{.8}{
	\begin{tabular}{ll}
		\toprule
Goal:& [restaurant]: \{informable: food international pricerange moderate area centre\} \{requestable: {\color{red}reference}\}\\

		\midrule
 User (Turn 0):&what restaurants in the centre serve international cuisine? \\
Golden R: &the [value\_name] and the [value\_name] serve [value\_food] food and are in the [value\_area] of town.\\
Golden BS: &[restaurant] food international area centre\\
Golden AS:& [restaurant] [inform] food area name       \\
 Mars-G R:& there are [value\_choice] restaurants that meet your criteria. do you have a price range in mind?\\
Mars-G BS: & [restaurant] food international area centre\\
  Mars-G AS:&   [restaurant] [inform] choice [request] price     \\
 
 	\midrule
User (Turn 1): &how about a place in the moderate price range?	\\
   Golden R: & [value\_choice] of the named restaurants are in the [value\_pricerange] price range.	\\
 Golden BS: &[restaurant] food international pricerange moderate area centre\\
 Golden AS:&   [restaurant] [inform] choice price   \\
  Mars-G R:  & there are [value\_choice] restaurants that meet your criteria. [value\_name] and [value\_name].\\& would you like me to book one of these for you? \\
  Mars-G BS: &[restaurant] food international pricerange moderate area centre \\
  Mars-G AS:&   [restaurant] [inform] choice name [offerbook]  \\
 
  	\midrule
 User (Turn 2):  & {\color{red}ok, can you book a table for 6 at 12:00 on tuesday at the varsity restaurant?} \\&{\color{red}i will need a reference number too, please.}	\\
  	Golden R: & certainly. i will have that reference number for you in just one second. \\
  	Golden BS: & [restaurant] food international pricerange moderate area centre \\
Golden AS:&    [restaurant] [offerbook]   \\
  	  Mars-G R: &  {\color{red}booking was successful. the table will be reserved for 15 minutes. reference number is:}\\& {\color{red}[value\_reference].} is there anything else i can help you with?\\
  	 Mars-G BS: & [restaurant] food international pricerange moderate area centre time 12:00 day tuesday people 6\\
  Mars-G AS:&     {\color{red}[restaurant] [offerbooked] reference} [general] [reqmore] \\

  	 	\midrule
  	 	User (Turn 3):   &thanks. looking forward to it.\\
  	 	  Golden R: & yes i can. {\color{red}i have booked you a table for [value\_time] on [value\_day] for [value\_people] people.} \\&{\color{red}your reference number is [value\_reference].} anything else i can help you with?\\
  	 	 	Golden BS:  &  [restaurant] food international pricerange moderate area centre time 12:00 day tuesday people 6 \\
  	 	 	Golden AS:&    [restaurant] {\color{red}[offerbooked]} time {\color{red}reference} people day \\
  	 	     Mars-G R: &you're welcome. have a great day!\\
  	 	 Mars-G BS:  &  [restaurant] food international pricerange moderate area centre time 12:00 day tuesday people 6\\
  Mars-G AS:&  [general] [welcome] [bye]    \\
    
		\bottomrule
	\end{tabular}}\caption{An example in dialog session sng0636. The requestable slot error is caused by automatic evaluation. \label{tab:case6}}
\end{table*}

\begin{table*}[ht]
  \centering
  \scalebox{.8}{
	\begin{tabular}{ll}
		\toprule
Goal:& [attraction]: \{informable: area north\} \{requestable: {\color{red}postcode phone} \}\\
		\midrule
 User (Turn 0):& i am looking for something to do in town. are there any interesting attraction in the north part of town?\\
Golden R: &yes , there are [value\_choice] places: i see a [value\_type], a [value\_type], and two [value\_type]. \\&would one of these work for you?\\
Golden BS: &[attraction] area north\\
Golden AS:&     [attraction] [inform] type choice [general] [reqmore]   \\
 Mars-G R:& there are [value\_choice] [value\_type] and [value\_type] in the [value\_area]. do you have a preference? \\
Mars-G BS: &[attraction] area north\\
  Mars-G AS:&    [attraction] [inform] area type choice [request] type    \\
 
 	\midrule
User (Turn 1): &ooh, a pool would be fun. {\color{red}can you give me their contact info?}	\\
   Golden R: & great! [value\_address] is located at [value\_address], kings hedges postcode is [value\_postcode] \\&and their phone is [value\_phone]. is there anything else i can help you with?	\\
 Golden BS: &  [attraction] area north	\\
 Golden AS:&     [attraction] [inform] name postcode phone address [general] [reqmore] \\
  Mars-G R:  & {\color{red}[value\_name] is located at [value\_address]. their phone number is [value\_phone].}\\
  Mars-G BS: &[attraction] area north type swimming pool \\
  Mars-G AS:&   {\color{red}[attraction] [inform] name phone address } \\

		\bottomrule
	\end{tabular}}\caption{An example in dialog session mul1811.  The requestable slot error is acceptable.\label{tab:case7}}
\end{table*}
\begin{table*}[ht]
  \centering
  \scalebox{.8}{
	\begin{tabular}{ll}
		\toprule
Goal:& [train]: \{informable: destination stansted airport  day  friday  arrive 15:30 departure cambridge\} \\
&  \;\;\;\;\;\;\;\;\;\;  \{requestable: {\color{red}reference}\}\\

		\midrule
 User (Turn 3):& i will also need a train leaving cambridge.\\
Golden R: &sure, what is your arrival information?\\
Golden BS: &[train] departure cambridge\\
Golden AS:&       [train] [request] destination \\
 Mars-G R:& what day will you be traveling, and where will you be traveling to?\\
Mars-G BS: &[train] departure cambridge\\
  Mars-G AS:&  [train] [request] destination day      \\
 
 	\midrule
User (Turn 4): &i need to go to stansted airport on friday.	\\
   Golden R: & we have [value\_choice] different departure times. any time table your looking at? 	\\
 Golden BS: & [train] destination stansted airport departure cambridge 	\\
 Golden AS:&     [train] [inform] choice [request] leave \\
  Mars-G R:  &there are [value\_choice] trains that meet your criteria. do you have a preferred departure or arrival time?\\
  Mars-G BS: &  [train] destination stansted airport day friday departure cambridge\\
  Mars-G AS:&  [train] [inform] choice [request] leave arrive   \\
 
  	\midrule
 User (Turn 5): & i need to be in stansted airport by 15:30 please.	\\
  	Golden R: & the soonest it will get you there is [value\_arrive] {\color{red}your reference number is} [value\_reference], anything else? \\
  	Golden BS: & [train] destination stansted airport arrive 15:30 departure cambridge people 1 \\
Golden AS:&  [train] {\color{red}[offerbooked] arrive reference} [general] [reqmore]   \\
  	  Mars-G R: & [value\_id] will arrive at [value\_arrive]. {\color{red}would you like me to book that for you?}\\
  	 Mars-G BS: &[train] destination stansted airport day friday arrive 15:30 departure cambridge \\
  Mars-G AS:&   [train] [inform] arrive id {\color{red}[offerbook] }  \\

		\bottomrule
	\end{tabular}}\caption{An example in dialog session pmul3247. The requestable slot error is acceptable. \label{tab:case8}}
\end{table*}
\begin{table*}[ht]
  \centering
  \scalebox{.8}{
	\begin{tabular}{ll}
		\toprule
Goal:& [attraction]: \{informable: type architecture area centre\} \{requestable: {\color{red}address phone postcode}\}\\

		\midrule
 User (Turn 0):& i want to find a place to go in the centre to enjoy some entertainment. can you help?\\
Golden R: &there are [value\_choice] attraction types in the [value\_area], did you have one in mind?\\
Golden BS: &[attraction] area centre\\
Golden AS:&   [attraction] [inform] area choice [request] name     \\
 Mars-G R:&i am sorry, there are no [value\_type] attractions in the [value\_area]. would you like to try another area? \\
Mars-G BS: &[attraction] area centre type entertainment\\
  Mars-G AS:&      [attraction] [nooffer] area type [request] area   \\
 
 	\midrule
User (Turn 1): &yes, i am looking for something with the type of architecture. thanks!\\
   Golden R: & i would recommend [value\_name] as a great place to see interesting [value\_type].	\\
 Golden BS: & [attraction] area centre type architecture	\\
 Golden AS:&    [attraction] [recommend] name type   \\
  Mars-G R:  &there are [value\_choice] [value\_type] attractions in the [value\_area]. i recommend [value\_name].\\
  Mars-G BS: & [attraction] area centre type architecture\\
  Mars-G AS:&   [attraction] [recommend] name [inform] type choice area  \\
 
  	\midrule
 User (Turn 2):  & okay, {\color{red}could you give me the address and phone number?}	\\
  	Golden R: & sure. the address is [value\_address]. the phone number is [value\_phone].\\& is there anything else i can help you with today? \\
  	Golden BS: & [attraction] area centre type architecture \\
Golden AS:&      [attraction] {\color{red}[inform] address phone} [general] [reqmore] \\
  	  Mars-G R: & the address is [value\_address] and the phone number is [value\_phone]. \\
  	 Mars-G BS: & [attraction] area centre type architecture\\
  Mars-G AS:&     [attraction] {\color{red}[inform] address phone}  \\

		\bottomrule
	\end{tabular}}\caption{An example in dialog session pmul1320. Noisy dialog annotations cause the requestable slot error. \label{tab:case9}}
\end{table*}

\end{document}